\documentclass{clv3}
\usepackage{hyperref}
\usepackage{xcolor}
\definecolor{darkblue}{rgb}{0, 0, 0.5}
\hypersetup{colorlinks=true,citecolor=darkblue, linkcolor=darkblue, urlcolor=darkblue}

\issue{1}{1}{2022}

\runningtitle{EDV and UD Parsing Performance}
\runningauthor{Anderson and G\'{o}mez-Rodr\'{i}guez}

\usepackage{amsmath}
\usepackage{amssymb}
\usepackage{url}
\usepackage{graphicx,float}
\usepackage{multicol,array,booktabs,multirow}
\usepackage{subfigure}
\usepackage{bm}
\usepackage{tikz-dependency}

\newcommand{\meanL}{$\langle L_{\textrm{test}}\rangle$}
\newcommand{\meanLtr}{$\langle L_{\textrm{train}}\rangle$}
\newcommand{\ccol}[1]{\multicolumn{1}{c}{\textbf{#1}}}

\definecolor{deppink}{HTML}{a92b7a}
\definecolor{depgreen}{HTML}{679033}
\definecolor{deporange}{HTML}{be6320}
\definecolor{depblue}{HTML}{165f77}
\definecolor{depred}{HTML}{b12019}
\definecolor{depgrey}{HTML}{4a4a53}
\definecolor{deppurple}{HTML}{4888ab}

\makeatletter
\g@addto@macro{\endtabular}{\rowfont{}}
\makeatother
\newcommand{\rowfonttype}{}
\newcommand{\rowfont}[1]{
\gdef\rowfonttype{#1}#1\ignorespaces%
}
\makeatother

\begin{document}
\title{The Impact of Edge Displacement Vaserstein Distance on UD Parsing Performance\thanks{Action editor: Kevin Duh. Submission received: 26 July 2021; revised version received: 11 January 2022; accepted for publication: 25 February 2022.}}
\author{Mark Anderson}
\affil{Universidade da Coru\~{n}a, CITIC\\m.anderson@udc.es}

\author{Carlos G\'{o}mez-Rodr\'{i}guez}
\affil{Universidade da Coru\~{n}a, CITIC\\carlos.gomez@udc.es}


\maketitle

\hphantom{-}

\begin{abstract}
We contribute to the discussion on parsing performance in NLP by introducing a measurement that evaluates the differences between the distributions of edge displacement (the directed distance of edges) seen in training and test data. We hypothesize that this measurement will be related to differences observed in parsing performance across treebanks. We motivate this by building upon previous work and then attempt to falsify this hypothesis by using a number of statistical methods. We establish that there is a statistical correlation between this measurement and parsing performance even when controlling for potential covariants. We then use this to establish a sampling technique that gives us an adversarial and complementary split. This gives an idea of the lower and upper bounds of parsing systems for a given treebank in lieu of freshly sampled data. In a broader sense, the methodology presented here can act as a reference for future correlation-based exploratory work in NLP.
\end{abstract}
\section{Introduction}
Evaluating the performance of NLP systems is an important task that is often  done using a well-established metric or set of metrics. Error analysis often just includes cherry-picking examples that are easy to discuss but don't necessarily give a clear picture of the quality of systems. However, in the context of syntactic parsing, plenty of literature has been written discussing what factors influence parsing performance and it is towards this discussion that this work contributes. We do so by looking at the edge displacement of nodes (the directed distance between the position of the node and its head, see Figure \ref{fig:tree_example}) and the corresponding distributions over samples. More specifically, we evaluate the distributions seen in training and test data of treebanks and use the Vaserstein distance to measure the difference between these two distributions. We then compare this with the parsing performance of two different systems that are broadly speaking a transition-based and graph-based parser.

\paragraph{Hypothesis} We postulate that the differences between the edge dependency displacement distributions of the training and test data of treebanks (as measured by the Vaserstein distance, formally introduced in Section 3.1) are related to the performance of parsers (as defined by the labeled attachment score). We use a number of methods in an attempt to falsify this hypothesis and conclude that based on the data and systems used in this analysis, it cannot be fully refuted. However, the sentence-length binning analysis tempers our complete confidence in this hypothesis.
\paragraph{Utility} We suggest using the observed correlation of Vaserstein distances between edge displacement distributions and parsing performance to guide a sampling method to create adversarial and complementary splits better suited for evaluating parsers.
\begin{figure}[t!]
\centering
\begin{dependency}[edge style={depgrey!92, thick},label style={fill=white},edge slant=7]
\begin{deptext}[column sep=0.6em,ampersand replacement=\^]
In \^ \textcolor{depred}{\textbf{der}}  \^ \textcolor{depred}{\textbf{Not}}  \^ \textcolor{deppurple}{\textbf{frisst}} \^ der \^ Teufel \^ \textcolor{deppurple}{\textbf{Fliegen}} \\
\end{deptext}
\depedge[edge height=0.5cm]{3}{1}{\textsc{case}}
\depedge[edge below,edge height=0.5cm,edge style={deppink!80, thick},label style={fill=deppink!15}]{3}{2}{\textsc{det}}
\depedge{4}{3}{\textsc{obl}}
\depedge[edge height=.45cm]{6}{5}{\textsc{det}}
\depedge[edge height=1.05cm,edge slant=5]{4}{6}{\textsc{nsubj}}
\depedge[edge below,edge height=0.5cm,edge style={depblue!90, thick},label style={fill=depblue!35}]{4}{7}{\textsc{obj}}
\deproot[edge unit distance=2.5ex]{4}{\textsc{root}}
\end{dependency}
\caption{Example tree highlighting dependency displacement for two nodes. \textcolor{depred}{\textbf{der}} at position 2 with its head  \textcolor{depred}{\textbf{Not}} at position 3 has a \textsc{det} edge (in magenta) with a dependency displacement of $2-3=-1$. Similarly,  \textcolor{deppurple}{\textbf{Fliegen}} at position 7 with its head \textcolor{deppurple}{\textbf{frisst}} at position 4 has an \textsc{obj} edge with a dependency displacement of $7-4=3$. English: \textit{When in need the devil eats flies.} 
}\label{fig:tree_example}
\end{figure}

\section{Related Work}\label{sec:related}
In this section we give a brief overview of previous work focused on explaining parsing performance and also focused on dependency distance.
\subsection{Analyzing parsing performance}\label{sec:performance}
An obvious and well-attested predictor of parsing performance is the amount of training data available, which is typically observed to be logarithmically related to parsing performance \cite{sagae-etal-2008-evaluating,falenska-cetinoglu-2017-lexicalized,strzyz-etal-2019-viable,dehouck-etal-2020-efficient}. The lengths of sentences have also been observed to impact parsing performance, with longer sentences being harder to parse than shorter sentences \cite{mcdonald-nivre-2011-analyzing}. In a similar vein, others have highlighted the effect dependency distance has on parsers, namely that longer dependencies tend to be harder to predict \cite{mcdonald-nivre-2011-analyzing,anderson-gomez-rodriguez-2020-inherent,falenska-etal-2020-integrating}. Edge direction entropy and word order freedom has also been shown to have a meaningful effect \cite{alicante2012a,rehbein2017,gulordava2015,gulordava2016}. Although this is not consistently observed across all data: \citet{chung2010factors} found that for Korean this is not so strongly related to parsing performance as other features of the language such as its pro-drop tendencies. \citet{alicante2012a} only found it impacted Italian constituency parsing, but not dependency parsing.  Part-of-speech bigram perplexity \cite{berdicevskis2018using}, entropy over trees \cite{corazza2013information}, the degree of non-projectivity \cite{mcdonald-satta-2007-complexity}, and morphological complexity \cite{dehouck-denis-2018-framework,coltekin-2020-verification} have also been presented as explanations or measurements for differences in parsing performance . 

Analyses also focus on comparisons between parsing paradigms and algorithms. Transition-based parsers often appear to struggle with longer distance relations more than graph-based parsers \cite{mcdonald-nivre-2011-analyzing,falenska-etal-2020-integrating}. However, \citet{kulmizev-etal-2019-deep} observed that the use of contextualized word embeddings off-set the typical issues associated with transition-based parsers. \citet{de2017} investigated the performance of the same transition-based algorithm using a neural network implementation and also a classical implementation, observing the same tendency for performance to decline as dependency distance increased. \citet{anderson-gomez-rodriguez-2020-inherent} found that the similarity of the inherent displacement distributions of algorithms to the distributions of treebanks was meaningfully correlated with parsing performance when accounting for sentence length for different transition-based algorithms. Beyond this, different frameworks and annotation schemes have been found to perform differently, often related to one or more of the metrics mentioned above \cite{kuebler2008a,matsuzaki2008comparative,bosco2010comparing,mille2012how,alicante2012a,pretkalnina2014constructions}.

Differences between training and test data have also been evaluated. \citet{zhang2009correlating} looked at certain metrics such as the rate of out-of-vocabulary tokens and unseen part-of-speech trigams and observed some correlation between these and parsing performance. However, the main focus in this area is on domain shifts between training and test data. Although this issue is not unique to parsing, there have been extensive results showing that domain shift can result in very steep drops in performance if the domains are very different \cite{gildea2001corpus,bosco2010comparing,plank2010dutch,foster2010cba}. More recently, \cite{sogaard-2020-languages} proposed the ratio of tree structures in the test data that did not occur in the training data as a predictor of parsing performance, but the results presented were found to be spurious once covariants were accounted for \cite{anderson-etal-2021-replicating}. 
Here we present a similar analysis but with a measurement which is not so restricted, based on dependency displacement distributions.

\subsection{Dependency distance}
Dependency distance is hypothesized to be constrained by working memory restrictions resulting in distances being minimized \cite{gibson2000dependency,liu2017}. This has been corroborated by numerous corpus-based analyses \cite{cancho2004,liu2008,liu2007,buch2006,futrell2015,temperley2018}. Although different languages appear to adhere to these restrictions to varying extents \cite{jiang2015,gildea2010}.  This relates to NLP parsing because if different languages or treebanks adhere to this constraint more or less than others, it could result in differences in the achievable performance of parsers. \citet{hudson2017} also highlighted that mean dependency distance varies significantly between treebanks, but added that the direction of dependencies could impact parsing difficulty as well. Different syntactic traits associated with parsing difficulty have been shown to be correlated with an increase in dependency length, for example free-order languages \cite{gulordava2015} and with an increase in non-projective dependencies \cite{ferrer2016,gomez2017}.

\citet{GomPoLR2017} hypothesized that transition-based parsers perform adequately because they are biased towards short dependencies. This was somewhat corroborated by \citet{eisner2010} who improved parser performance by imposing limits on dependency length and using dependency lengths as a feature for their system. It was further substantiated by the work of \citet{anderson-gomez-rodriguez-2020-inherent} as described in Section \ref{sec:performance}.

The work presented here can be considered an extension of the previous work cited above where we use a method based on edge displacement distributions to compare differences between training and test data to attempt to explain variation in parsing performance across different treebanks.
\section{Methodology}
In this section we introduce the core principles behind the measurement we focus on in this article and we give the details of the parsing systems and data used in our analysis.
\subsection[EDV]{Edge displacement Vaserstein \\distance}\label{sec:edv}
We follow \citet{anderson-gomez-rodriguez-2020-inherent} and use edge displacement instead of distance as this gives us a measurement that encodes both distance and direction.  Fundamentally it is the signed distance of a node with respect to its head. We alter the definition from \citet{anderson-gomez-rodriguez-2020-inherent}, so that it better resembles the standard definition of physical displacement, that is the endpoint minus the starting point:
\begin{equation}\label{eqn:displacement}
    s_{\textrm{edge}} = x_{\textrm{node}} - x_{\textrm{head}}
\end{equation}
Then for a given treebank the edge displacement for each node is measured, excluding the root node and its displacement with respect to the dummy root as position 0. The range $[-30,30]$ is used for the distributions so that the measurement isn't impacted by potential unreliable long tails. This range covers 99.40\% of the edges in UD v2.5 and 99.38\% in UD v2.6. The distribution of edge displacements is then normalized such that it takes the form of a probability distribution. In this way, a probability distribution over 
displacements
is obtained for the training treebank and test treebank for each dataset. 
We then use these two probability distributions to calculate the Vaserstein distance \cite{vaserstein1969markov}, thus obtaining the edge displacement Vaserstein distance (EDV) for a given dataset. The Vaserstein distance (technically the Vaserstein-1 distance) is defined as follows \cite{vaserstein1969markov}:
\begin{equation}
    \ell(\mu,\nu) = \underset{\gamma\in\Gamma(\mu,\nu)}{\textrm{inf}}\int|x-y| d\gamma(x,y)
\end{equation}
where $\mu$ and $\nu$ are probability distributions of two random variables (in our case, the variables will correspond to dependency displacements), $x$ and $y$ are points in the x-axis of these probability distributions (i.e., concrete values of each of the variables), $|x-y|$ is the distance between two such values, and the infimum is with respect to $\gamma$, a coupling from $\Gamma$ which is the set of all joint distributions whose marginals are $\mu$ and $\nu$.

A more grounded interpretation of the Vaserstein distance is that it gives a measurement of how much \textit{mass} needs to be moved from each $x$ to each $y$ so that $\mu$ is transformed into $\nu$. As such, this metric is also known as earth mover's distance in computing science. Ultimately it gives a measurement of how different two distributions are, with larger values indicating a greater divergence and values approaching zero indicating similar distributions.

\begin{figure}[tb!]
    \centering
    \includegraphics[width=0.5\linewidth]{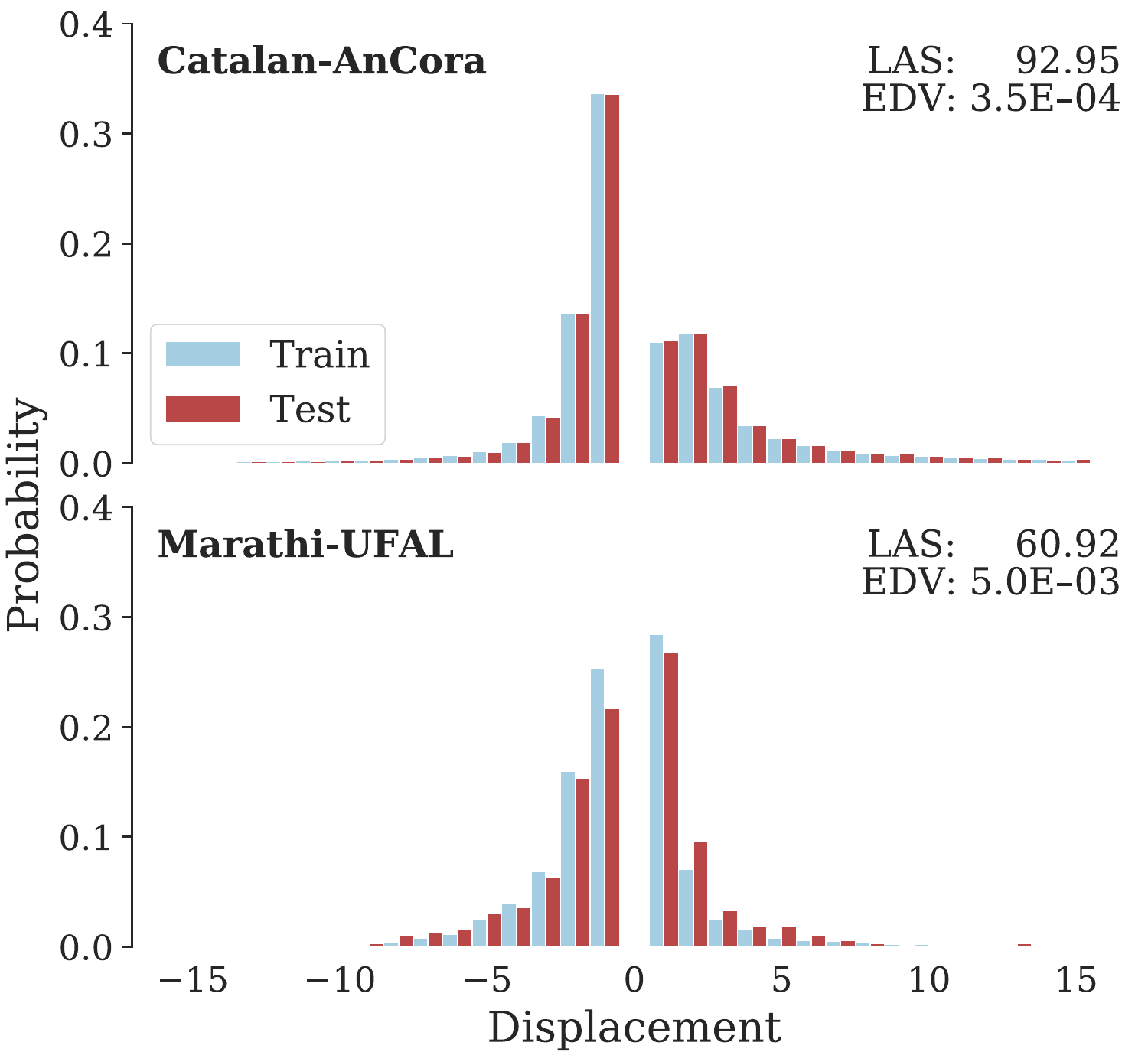}
    \caption{Example displacement distributions of the training and test data for Catalan-AnCora (top) and Marathi-UFAL (bottom) which exhibit the smallest and largest measured EDV values in UD v2.6. While both EDV values are small, there is an order of magnitude difference between them. LAS is shown for UDPipe 2.0.}
    \label{fig:distros}
\end{figure}

\paragraph{Example} Distributions are shown for two treebanks from the Universal Dependency (UD) v2.6 treebanks in Figure \ref{fig:distros}. As can be seen, Catalan-AnCora has very similar distributions for its training and test data, which is reflected in a small EDV of $3\times10^{-4}$. Marathi-UFAL is also shown where differences between the two sets can be clearly seen despite the distributions following similar trends. This still results in a small EDV of $5\times10^{-3}$, but it is an order of magnitude greater than that observed for Catalan-AnCora. These two treebanks show the highest EDV (Marathi-UFAL) and the lowest (Catalan-AnCora) and so show the range of EDV values observed in the data (the mean EDV observed in UD v2.6 is $1.40 (0.85)\times10^{-3}$, and $1.35 (0.87)\times10^{-3}$ for UD v2.5) . Despite the values of EDV both being fairly small, there is a large difference in performance seen for these two treebanks with Catalan-AnCora achieving a labeled attachment score (LAS) of 92.95 when using UDPipe 2.0, and Marathi-UFAL only achieving 60.92. There are clearly other contributing factors relating to the difference in performance between these two treebanks (not least training data size, as Marathi-UFAL only has 373 training instances whereas Catalan-AnCora has 13,123) which we have discussed above and which we take into consideration in our analysis discussed below.

\subsection{Parser systems}\label{sec:parsers}
We used two neural based parsers: version 1.2.1-devel (1.2) of UDPipe, and version 2.0 \cite{straka2017tokenizing,straka2018udpipe}. For UDPipe 1.2 we use models 2.5\footnote{\href{https://lindat.mff.cuni.cz/repository/xmlui/handle/11234/1-3131}{https://lindat.mff.cuni.cz/repository/xmlui/handle/11234/1-3131}} and for UDPipe 2.0 we use models 2.6.\footnote{\href{https://lindat.mff.cuni.cz/services/udpipe/}{https://lindat.mff.cuni.cz/services/udpipe/}} We opted to use these systems as the models have been optimized for their respective UD treebank and UDPipe 1.2 is a transition-based system while UDPipe 2.0 is a graph-based system, thus allowing us to evaluate EDV for different parser systems. Furthermore, UDPipe 1.2 came 8th out of 33 at the CoNLL 2017 shared task and was used as the baseline model for comparison of systems submitted to the CoNLL 2018 shared task, where it came 18th out of 26 with respect to average LAS. For its part, UDPipe 2.0 was one of the top performing parsers of the 2018 shared task, tied for the 3rd place \cite{zeman-EtAl_conll17,zeman-conll18}. An earlier version of UDPipe 2.0 was also one of the leading systems at the SIGMORPHON 2019 shared task, and the winner of EvaLatin 2020 \cite{mccarthy-etal-2019-sigmorphon,sprugnoli-etal-2020-overview}. 

Both systems include tokenization and sentence segmentation capabilities, but we fed gold tokenized data to the systems as we are interested in the impact of EDV on parsing specifically and not how it relates to these preliminary tasks. When running the systems, we opted to run the taggers when parsing so as to use the systems close to how they were intended to be used, even though we are not interested in the tagging performance (of UPOS and mfeats). This results in using predicted tags at runtime.  

\textbf{UDPipe 1.2} is a basic feed-forward neural transition-based parser which uses a simple feature function as input for each timestep \cite{chen2014,udparsing_straka15}. We used models 2.5 which were pre-trained on UD v2.5 treebanks, resulting in 94 parsers on separate treebanks. Each model is optimized for each treebank which includes the type of algorithm and oracle used. Details of the system can be found in \citet{udparsing_straka15} and \citet{straka2016udpipe}.

\textbf{UDPipe 2.0} is based on the graph-based biaffine parser of \citet{dozat20161} where the hidden representations of tokens from BiLSTM layers are mapped into two separate perceptron layers, considered representations of the tokens as a head and as a dependent, which are combined using a biaffine attention mechanism, resulting in a probability distribution over all other tokens in a sentence indicating the probability that any given token is its head. A well-formed tree is then enforced using the Chu–Liu/Edmonds' algorithm \cite{chuliu65,edmonds1967optimum}. We could not run Czech-PDT, Hindi-HDTB, German-HDT, and Russian-SynTagRus as the website had issues with large files, so we ended up with results from 90 models. Note that while the treebanks used for UDPipe 1.2 and 2.0 are very similar, they are not exactly the same. There are few differences in the actual treebanks included and there are also differences within given treebanks between iterations of UD releases.
\paragraph{Data}
We used UD treebanks for our analysis (as such, we lay no claim to any results that span different frameworks). We used the sets of treebanks that correspond to the parser models we used for each system, namely UD v2.6 with UDPipe 2.0 and UD v2.5 for UDPipe 1.2. We also used UD v2.7 to extend our analysis beyond the pretained model for evaluation of the linear regression model using unseen data. We picked treebanks that had no UDPipe 1.2 model but contained both training and test data and which contained at least 100 sentences in the training data. We also used UD v2.7 for a proof of concept for using EDV to guide sampling for a more robust evaluation procedure for parsers. This resulted in 94 treebanks for UDPipe 1.2, 90 treebanks for UDPipe 2.0, 11 treebanks for evaluating the UDPipe 1.2 linear model, and 105 treebanks for the sampling work.
\subsection{Statistical methods}\label{sec:stats}
\begin{figure}[t!]
    \centering
    \includegraphics[width=0.49\linewidth]{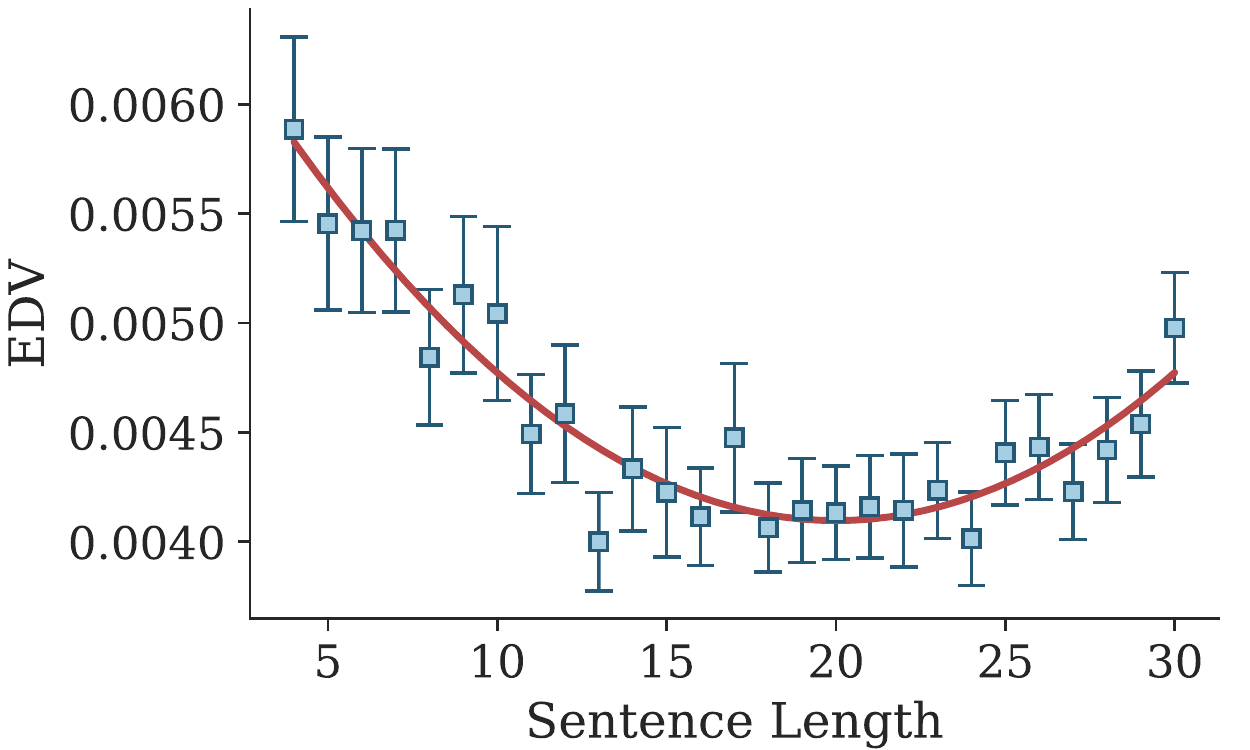}
    \caption{EDV between sub-samples of the training and test data binned by sentence length for UD v2.6 (111 treebanks).}
    \label{fig:edda_vs_length}
\end{figure}

The statistical analysis was undertaken using the Pingouin Python library version 0.3.8, except the partial coefficients (\S\ref{sec:partial}) were calculated using version 0.4.0 which corrected errors associated with these \cite{vallat2018pingouin}.
\paragraph{Correlation coefficients}
We evaluate the impact variables have on parsing performance by measuring their correlation coefficients with respect to LAS. We use a non-parametric correlation coefficient in the form of Spearman's $\rho$, which measures the correlation between variables and assesses the monotonic relationship between them. We do not use Pearson's r as the data being analyzed do not strictly adhere to bivariate probability distributions and the sample sizes are small enough that this can affect the measurement's sensitivity. Further, Pearson's r is less robust with respect to outliers. For each coefficient, we report the correlations and the corresponding p-value. For the main correlation results, we include the upper and lower bounds of the 95\% confidence interval, the coefficient squared (a measure of the proportion of explained variance), the adjusted coefficient which somewhat tempers the coefficient's bias, and the power of the analysis. For p-values, we report the exact value unless the value is less than 0.001, following common practice \cite{apa:6}.
\paragraph{Partial correlations} We make use of partial correlations 
to evaluate the impact of covariants. This allows us to remove the impact of variables that are correlated with the control variable and the target variable, so as to avoid situations where a measurement seemingly explains X variance in the data but in reality it is merely a measurement of one or more basic variables.
\paragraph{Background removal} Here we take a standard method found in physics used to remove known background functions from data, for example removing the spectra associated with amorphous radiators from those associated with lattice-structure radiators to obtain enhanced spectra, that is without noise \cite{timm1969coherent}. Here we consider the variations associated with covariants as similar background data to be removed, so as to observe if there is any variation associated with EDV. Similar to partial correlations, removing the background signal of a potential covariant allows us to visually evaluate the specific impact a variable of interest has on the target variable. This involves fitting the control data and the target (e.g., the size of training data and LAS) and then dividing the target variable by the predicted values from this fit. This \textit{normalized} data is then used to fit a second potential covariant which too is used to divide the normalized target variable values. This can be repeated for any number of covariants. Ultimately a normalized version of the target variable is left and the control target of interest (e.g., EDV) is evaluated against these values and if a trend is still observed, it is evidence that this variable has an impact on the target variable even with the variance associated with these covariants removed. This technique ultimately acts as a way of tempering correlations we calculate and gives us a means of disentangling contributions that might not be caught by partial correlation calculations.
\paragraph{Linear regression} The preceding methods allow us to hone in on the impact of a given variable, but with linear regression we can fit models to the data with more than one variable. This allows us to evaluate the impact certain variables have when used with other covariants. For linear regression models we report the adjusted R$^2$ (the square of the residuals) as a measurement of the proportion of explained variance, which it equals when the residual mean is normalized so as to equal zero (as is the case in this analysis). In addition, we report the relative importance of each variable and the corresponding p-values \cite{sen1981introduction,groemping2006relative}.  
\paragraph{Sentence length binning} \citet{ferrer-i-cancho2014the} highlighted the impact mixing sentence lengths can have on treebanks analyses and \citet{anderson-gomez-rodriguez-2020-inherent} observed sentence-length dependencies when evaluating edge displacement distributions of treebanks and the inherent distributions of transition-based parsers. Considering this potential impact, we also undertake a sentence-length binned analysis. This simply entails constructing samples of each treebank based on the length of the sentences. We take bins ranging from 3 tokens to 30 tokens, as any shorter and the EDV has little meaning (i.e., with 2 tokens, there can only be one edge which can either be -1 or 1) and any longer and the number of instances in a given bin for a given treebank is too small to obtain a meaningful measurement. Note that parsers were trained on the full data and the binning procedure is undertaken solely at the analysis stage. Figure \ref{fig:edda_vs_length} shows the EDV calculated between training and test data for each sentence length bin for UD v2.6 (the corresponding data for UD v2.5 is shown in Figure \ref{fig:edda_vs_length_25} in \hyperref[appendix:vis]{Appendix~A}). It is clear that EDV does vary based on sentence length, but it remains to be seen whether that variation has an impact on parsing performance.
\paragraph{Variables assessed} Beyond assessing EDV and how it correlates to parsing performance (as given by LAS) we look at a number of variables which are potential covariants. First we look at the size of the training data (measured both in tokens and sentences), which as described above has been shown to correlate to parsing performance and could feasibly impact EDV measurements, that is larger treebanks allow for a more \textit{accurate} representation of a language's true underlying distributions of edge displacements so deviations with respect to the test data could be minimized, and vice versa: if the sample is too small, it could be some random sample at the fringes of what would be a standard distribution for a given language.
Similarly, we also consider the number of tokens and sentences in the test data. We also look at the mean sentence length of the test data, \meanL, as this theoretically puts a limit on the potential distribution of edge displacements and has been observed to impact parsing performance (i.e., longer sentences are harder to parse than shorter ones). For the sake of completeness, we also look at the mean length of the training data, \meanLtr. Finally, we look at the Vaserstein distance between the training and test distributions of sentence lengths (SLV) because it is feasible that EDV merely vaguely measures differences with respect to sentence length.

\section{Analysis and results}
In this section we describe the analysis in detail and discuss the results we obtained.
\subsection{Evaluating normality}
\begin{figure}[tb!]
    \centering
    \includegraphics[width=0.49\linewidth]{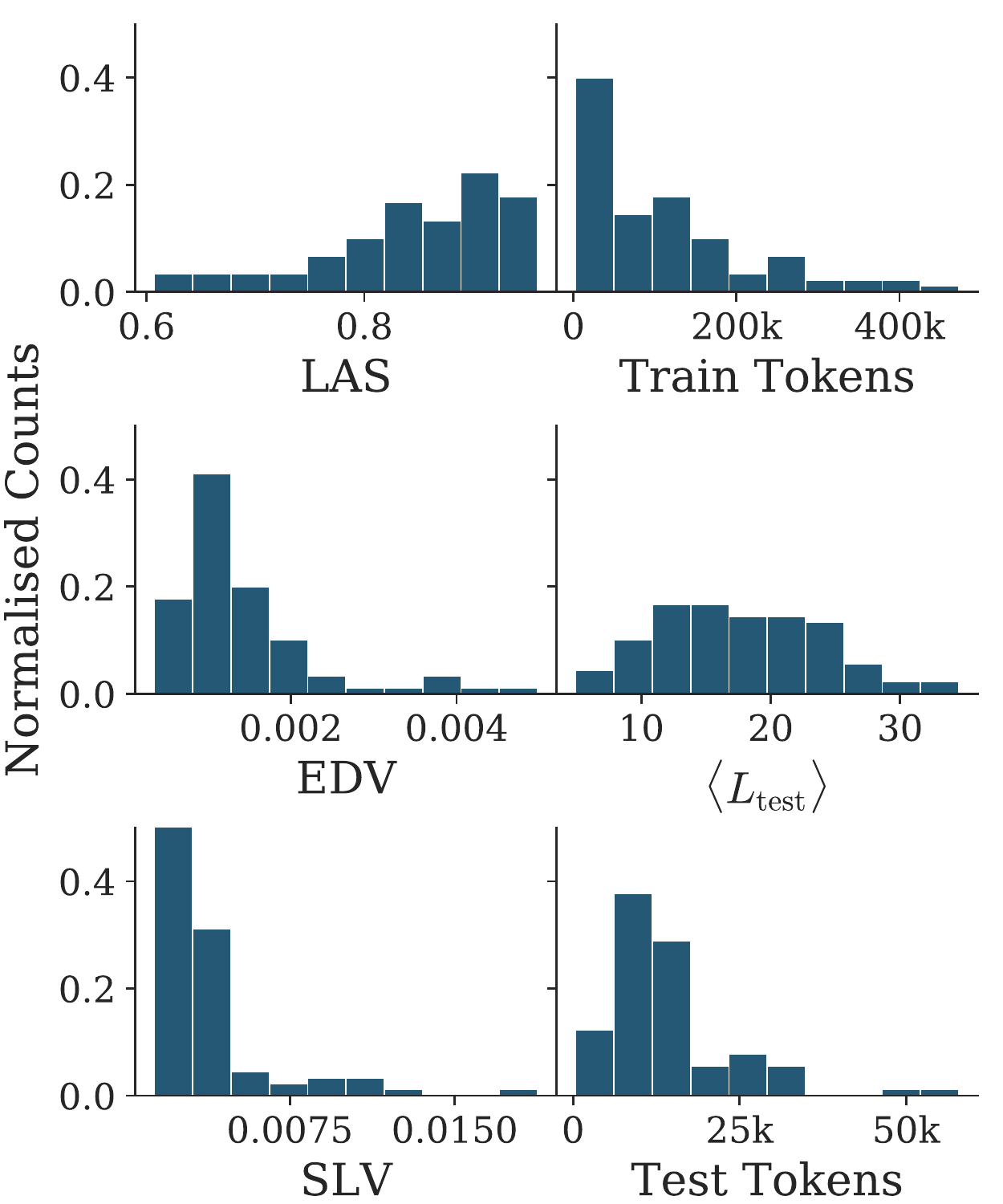}
    \caption{Distributions of the variables of interest in UD v2.6 (90 treebanks) in order to evaluate whether they are sampled from normal distributions.}
    \label{fig:not_normal}
\end{figure}

Here we justify the use of Spearman's $\rho$ for the following analysis. Figure \ref{fig:not_normal} shows the distributions of the variables of interest in our analysis (as described in Section \ref{sec:stats}) for UD v2.6 (the corresponding distributions for UD v2.5 are shown in Figure \ref{fig:not_normal_25} in \hyperref[appendix:vis]{Appendix~A}). Visually, it is clear that only \meanL{} could be sampled from a normal distribution.  

\begin{table}[b!]
    \centering
    \caption{Shapiro--Wilk tests to evaluate if samples are drawn from normal distributions for UD v2.5 (top) and UD v2.6 (bottom). Only the \meanL{} test has values for which the null hypothesis (i.e., normal distribution) cannot be rejected under any reasonable thresholds.}
    \label{tab:wilks}
    \begin{tabular}{lcrc}
    \toprule
    \textbf{Variable} & \textbf{W} & \ccol{p-value} & \textbf{Normal}  \\
    \midrule
    LAS & 0.920 & $<$0.001 & False \\
    Train Tokens & 0.418 & $<$0.001 & False \\
    EDV & 0.785 & $<$0.001 & False \\
    \meanL & 0.978 & 0.121 & \textbf{True} \\
    SLV & 0.686 & $<$0.001 & False \\
    Test Tokens & 0.350 & $<$0.001 & False \\
    \midrule
    LAS & 0.894 & $<$0.001 & False \\
    Train Tokens & 0.851 & $<$0.001 & False \\
    EDV & 0.761 & $<$0.001 & False \\
    \meanL & 0.985 & 0.402 & \textbf{True} \\
    SLV & 0.665 & $<$0.001 & False \\
    Test Tokens & 0.825 & $<$0.001 & False \\
    \bottomrule
    \end{tabular}
    
\end{table}
\begin{figure*}
    \centering
    \includegraphics[width=\linewidth]{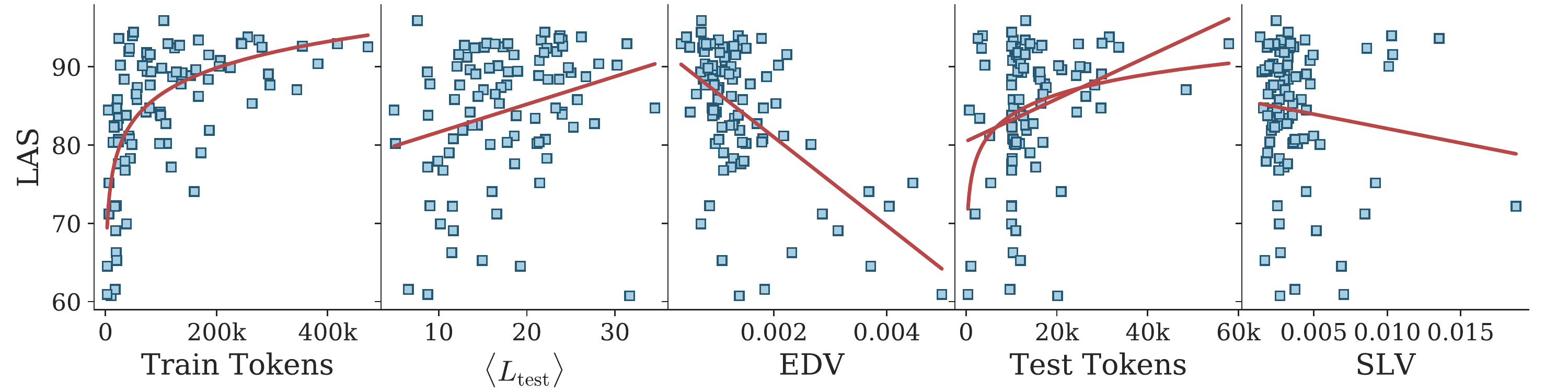}
    \caption{Visualization of LAS (for UDPipe 2.0 and UD v2.6) with respect to variables of interest with fits shown in red to highlight whether the data appears correlated or not.}
    \label{fig:correlations_wrt_las}
\end{figure*}
To thoroughly evaluate the variables for normality, we use the Shapiro--Wilk test \cite{shapiro1965an} as it is a higher power test compared to the alternatives, making it the most suitable for our fairly small sample size \cite{yap2011comparisons}. The values from the tests (W) and the corresponding p-values (where the null hypothesis is that the sample \textit{is} from a normal distribution) are shown in Table \ref{tab:wilks} for both UD v2.5 (top) and UD v2.6 (bottom). A smaller W indicates that a sample is not drawn from a normal distribution, but the more informative metric here is the p-value (as W is non-linear and difficult to interpret). Basically, larger p-values mean we cannot reject the null hypothesis that the sample is drawn from a normal distribution. Only \meanL{} has a large p-value and does so for both datasets (0.121 for UD v2.5 and 0.402 for UD v2.6). The left most column of Table \ref{tab:wilks} shows the result of the test based on the ever arbitrary distinction of significance, that is p-value $<0.05$. We are not particularly interested if one variable is or is not normally distributed, the important result here is that most variables including the control variable of interest (EDV) and the target variable (LAS) quite definitively do not follow normal distributions. This along with the other considerations mentioned in Section \ref{sec:stats} thoroughly justifies the use of Spearman's $\rho$. Further, it is useful that this coefficient doesn't specifically evaluate the linearity of relationships because not all variables assessed here are linearly related to parsing difficulty, but \textit{are} monotonically related. 

\subsection{Correlation coefficients}\label{sec:corrs}
Here we evaluate basic coefficients between the control variables and LAS and also between the potential covariants and EDV.
\subsubsection{Basic coefficients}

Figure \ref{fig:correlations_wrt_las} shows LAS against the control variables of interest for UDPipe 2.0 (the corresponding visualization for UDPipe 1.2 is shown in Figure \ref{fig:correlations_wrt_las_p1} in \hyperref[appendix:vis]{Appendix~A}). In the first subplot, it is fairly clear that LAS increases logarithmically with respect to the number of tokens in the training data, which corroborates the findings discussed above in Section \ref{sec:related}. It appears that the number of tokens in the test data is not associated with parsing performance for UDPipe 2.0, however, there is a potentially logarithmic relationship seen for UDPipe 1.2, but that could easily be down to a few serendipitously placed outliers. \meanL{} is loosely linearly related to LAS, but EDV seems like it is more strongly linearly related. SLV doesn't seem to be related to LAS, but there are a few clusters which upset the fitting procedure that should not affect the calculation of the corresponding Spearman $\rho$ for this relation. Note that we do not visualize all variables for the sake of space and to avoid redundancy, that is the number of training tokens is more strongly correlated to parsing performance than the number of training sentences (as seen in Table \ref{tab:las_spearmans}). 
\begin{table}[t!]
    \centering
    \small
     \caption{Spearman's $\rho$ for correlations between variables of interest and LAS.}
    \label{tab:las_spearmans}
    \begin{tabular}{llrr}
    \toprule
   \textbf{Parser} &\textbf{Variable}  & \ccol{$\rho$} & \ccol{p-value} \\ \midrule
        \multirow{8}{*}{UDPipe 1.2} & Train Tokens &  0.660 & $<$0.001  \\
        &Train Trees   & 0.535 & $<$0.001 \\
        &\meanLtr &  0.376 & $<$0.001 \\ 
        &Test Tokens  & 0.433 & $<$0.001  \\
        &Test Trees   & 0.208 & 0.045 \\
        &\meanL &  0.351 & 0.001 \\ 
        &SLV &  -0.191 & 0.065 \\
        &EDV  &  -0.492 & $<$0.001 \\
        \midrule
        \multirow{8}{*}{UDPipe 2.0} & Train Tokens &   0.605 & $<$0.001\\
        &Train Trees  &  0.467 & $<$0.001 \\
        &\meanLtr &  0.323 & 0.002 \\
        &Test Tokens  & 0.309 & 0.003\\
        &Test Trees  & 0.073 & 0.496 \\
        &\meanL &  0.309 & 0.003 \\
        &SLV &  -0.086 & 0.422 \\
        &EDV &  -0.466 & $<$0.001 \\ 
      \bottomrule
    \end{tabular}
   
\end{table}

Table \ref{tab:las_spearmans} shows the corresponding Spearman $\rho$ values for the data shown in Figures \ref{fig:correlations_wrt_las} and \ref{fig:correlations_wrt_las_p1} and the remaining variables mentioned above in Section \ref{sec:stats}, that is control variables related to LAS. First, we want to note that measuring data in tokens rather than sentences results in stronger correlations for both test and training and for both parsers (with the number of test instances not even being correlated to LAS for UDPipe 2.0). Based on this, we use the number of tokens in the training and test data from here on in. Also, the number of training tokens is the variable most strongly correlated with parsing performance, but the next strongest for both systems (excluding the number of training sentences) is actually EDV. SLV is not correlated at all for UDPipe 2.0 and only weakly so for UDPipe 1.2, with a p-value higher than any arbitrary threshold of significance.  

\begin{figure}[b]
    \centering
    \includegraphics[width=0.92\linewidth]{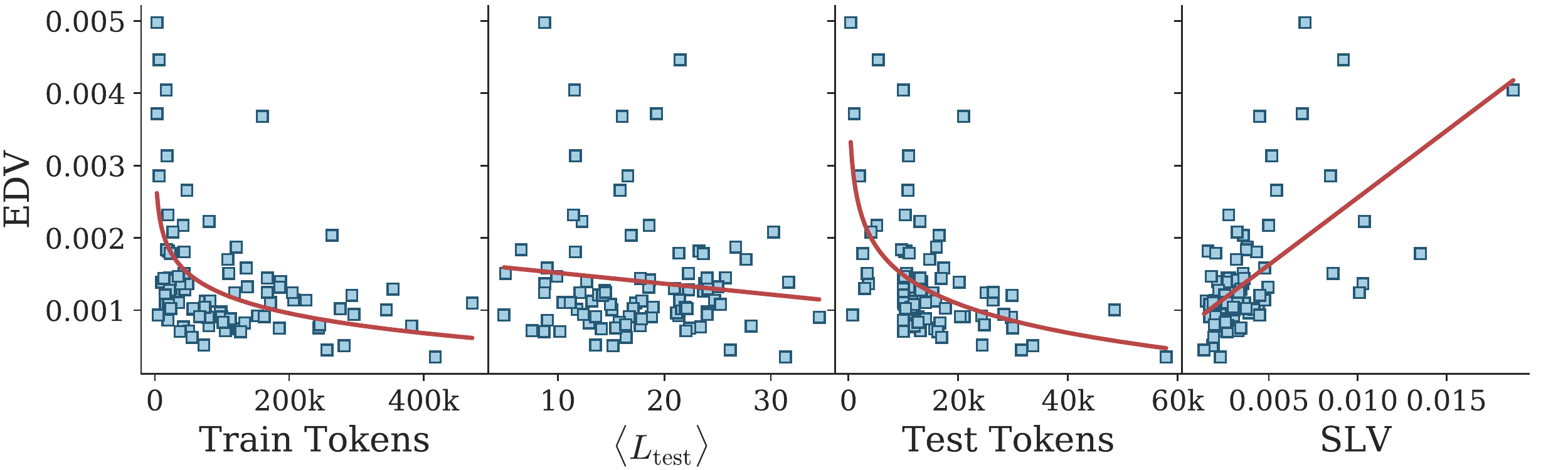}
    \caption{Visualization of EDV (for UD v2.6) with respect to variables of interest with fits shown in red to highlight whether the data appears correlated or not.}
    \label{fig:edv_vs_others}
\end{figure}

Next we investigate how the variables most strongly correlated to LAS correlate with one another, that is we check for potential covariants. Figure \ref{fig:edv_vs_others} shows how pertinent variables relate to EDV. Clearly, the number of tokens in the training data and the test data are strongly related and, as one would expect, SLV looks related (confirmed by the actual correlation coefficient of 0.549 with a p-value less than 0.001 as seen in Table \ref{tab:inter_spearmans}). However, as SLV is not correlated to parsing performance, it is not necessary to consider it when evaluating EDV with respect to LAS. It seems like \meanL{} is not clearly related to EDV despite our expectations that it would be. 
\begin{table}[t!]
\centering
    \small
     \caption{Spearman's $\rho$ for different pairs of variables.}
    \label{tab:inter_spearmans}
    \begin{tabular}{llrr}
    \toprule
     \textbf{Parser} & \ccol{Variables} & \ccol{$\rho$} & \ccol{p-value} \\ \midrule
      \multirow{7}{*}{UDPipe 1.2} & Train Tokens --- EDV & -0.480 & $<$0.001 \\
      & \meanL --- EDV & -0.080 & 0.443 \\
       & \meanLtr --- EDV & -0.089 & 0.393 \\
      & Test Tokens --- EDV & -0.523 & $<$0.001 \\
      & SLV --- EDV &0.617 & $<$0.001 \\
      & Test --- Train (Tokens) & 0.772 & $<$0.001\\
      & \meanL --- Train Tokens & 0.149 & 0.153 \\
       \midrule
        \multirow{7}{*}{UDPipe 2.0} & Train Tokens --- EDV & -0.424 & $<$0.001 \\
        & \meanL --- EDV & -0.025 & 0.817 \\
        & \meanLtr --- EDV & -0.023 & 0.833 \\
        & Test Tokens --- EDV & -0.446 & $<$0.001 \\
        & SLV --- EDV & 0.549 & $<$0.001\\
        & Test --- Train (Tokens) & 0.659 & $<$0.001 \\
         & \meanL --- Train Tokens & 0.096 & 0.370 \\
        \bottomrule
    \end{tabular}
   
\end{table}

The corresponding correlations are found in Table \ref{tab:inter_spearmans} alongside correlations between other variables as well. The correlations clearly corroborate the trends observed in Figure \ref{fig:edv_vs_others}. \meanLtr{} is not shown in Figure \ref{fig:edv_vs_others} but it behaves similarly to \meanL{}, closely echoing the measured correlations between \meanL{} and EDV for both systems.
We also show the correlation between the number of training tokens and test tokens as typically the amount of data for both are linked (i.e., it is not particularly common for a treebank to have a huge training set but a tiny test set, although the opposite does occur, eg. Kazakh-KTB). For both sets of data the correlations are high (0.772 for UD v2.5 and 0.659 for UD v2.6) both with p-values below 0.001. We assume, therefore, that these measurements loosely capture the same aspect of treebanks and use the number of training tokens as the best option: it is more strongly correlated to LAS by a large amount and is similarly correlated to EDV if slightly less so than the number of test tokens. We further justify this choice in Section \ref{sec:bcg}. Lastly, we show the correlation of \meanL and the number of training tokens as it has been noted that smaller treebanks (especially very low-resource treebanks) not only have less training instances but also sentences tend to be shorter \cite{dehouck2020data}. However, we don't find any correlation in these datasets, presumably because this issue is not prevalent once a certain threshold of data size is reached.

\begin{figure}[t!]
    \centering
    \includegraphics[width=0.85\linewidth]{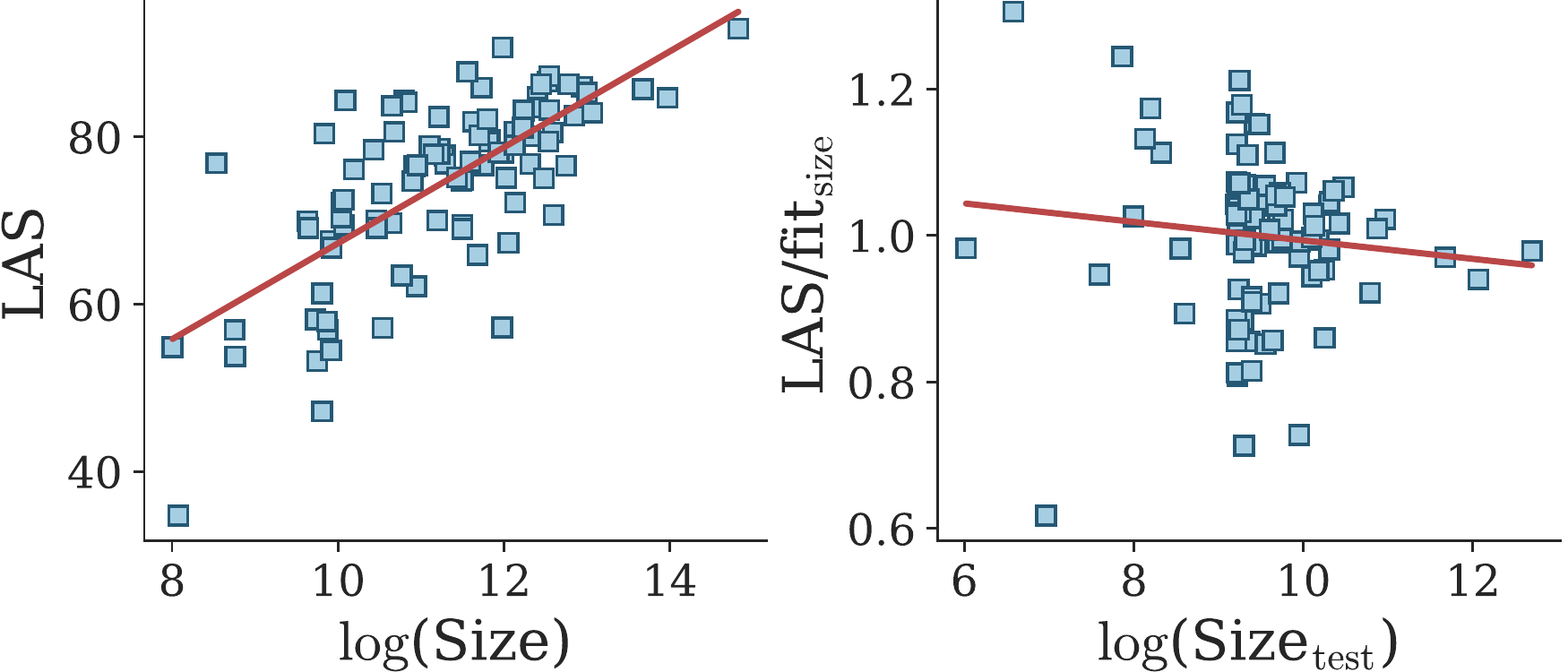}
    \caption{Background removing method used to evaluate whether the number of test tokens carries additional information with respect to the number of training tokens for UDPipe 1.2 and UD v2.5. Correlation between the number of test tokens and LAS  is 0.433 (p-value$<$0.001) and that between the number of test tokens and the normalized LAS (right plot) is -0.123 (p-value$=$0.236).
}
    \label{fig:bcg_test25}
\end{figure}

\subsubsection{Background removal}\label{sec:bcg}
As described above, we removed the background signal associated with other variables to evaluate the independent relationship of certain variables. First, we evaluated whether the number of test tokens actually captured a different aspect of the treebanks with respect to parsing performance. Figure \ref{fig:bcg_test25} shows this process for UDPipe 1.2, where the first plot shows LAS against the number of training tokens and the second plot shows the normalized LAS (LAS / fit from first plot) against test tokens.

We show this process for UDPipe 1.2 rather than 2.0 which we have used for the visual representations in the main body thus far (the corresponding plot for UDPipe 2.0 is shown in Figure \ref{fig:bcg_test} in \hyperref[appendix:vis]{Appendix~A}) as the visual relationship observed for UDPipe 1.2 between the number of test tokens and LAS was much more convincing than for UDPipe 2.0 and the correlation reported in Table \ref{tab:inter_spearmans} was higher for UDPipe 1.2. It is clear that once removing the signal associated with the number of training tokens, the signal associated with the number of test tokens disappears. This is backed up by the correlations observed for the number of test tokens and LAS (0.433, p-value$<$0.001) disappearing when comparing the number of training tokens to the normalized LAS with a correlation of $-$0.123 (p-value$=$0.236). 

We note here that when looking at the partial coefficient for the number of test tokens for UDPipe 1.2 when using the number of training tokens as a covariant, we obtain a coefficient of -0.325 (p-value$=$0.001) which is not particularly meaningful and highlights the fragility of correlation coefficients. In fact, the reversal of the sign is indicative of multicollinearity, exactly what we anticipated these variables to be \cite{farrar2011multicollinearity}. For UDPipe 2.0 the same partial correlation is $-$0.045 (p-value=0.671) and so it is even clearer for this system. 
\begin{figure}[!t]
    \centering
    \includegraphics[width=\linewidth]{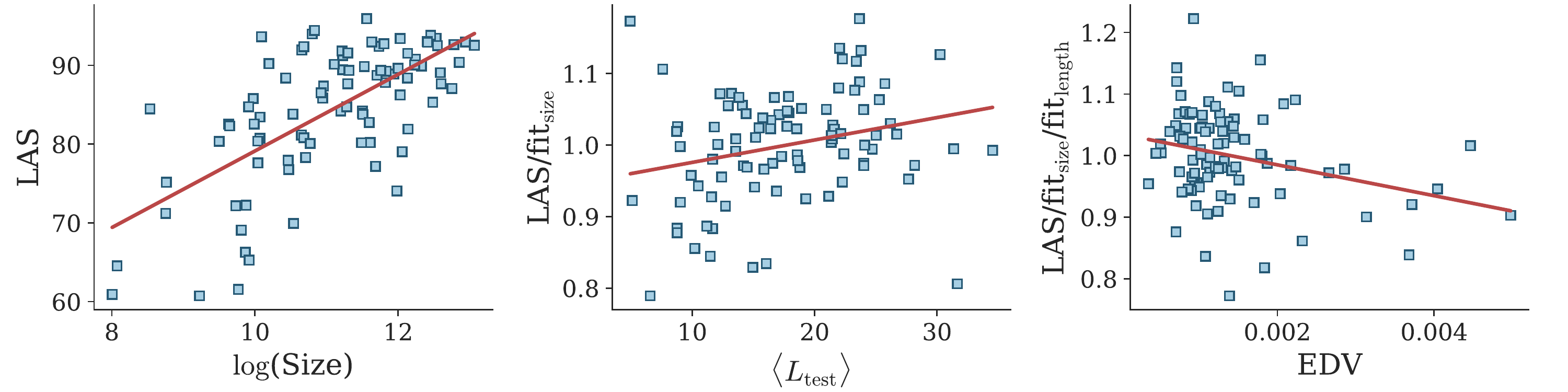}
    \caption{Background removal method to evaluate whether a correlation is observed between EDV and LAS (for UDPipe 2.0 and UD v2.6) after removing the variation associated with the training test size and \meanL. The correlation between EDV and LAS is -0.466 (p-value$<$0.001), the correlation between EDV and the LAS normalized by the variance associated with number of tokens in training data is -0.222 (p-value$=$0.036), and the correlation for the fully normalized LAS (removing the variance associated with \meanL) is -0.283 (p-value$=$0.007).
}
    \label{fig:noise-signal}
\end{figure}
\begin{table}[b!]
    \centering
        \footnotesize
        \caption{Partial coefficients (except for rows with None in Covariant(s) column) for EDV with respect to LAS for UDPipe 1.2 and UD v2.5 (top) and for UDPipe 2.0 and UD v2.6 (bottom). Shown is the coefficient itself ($\rho$), the 95\% confidence interval (CI95\%), $\rho^2$ as an indication of the proportion of explained variance, the adjusted $\rho^2$ (Adj. $\rho^2$) as a less biased version of $\rho^2$, the corresponding p-values, and the achieved power of the test (power).}
    \label{tab:re_results}
    \begin{tabular}{llrrrcrr}
    \toprule
\textbf{Parser} & \textbf{Covariant(s)} &  \ccol{$\bm{\rho}$} &  \ccol{CI95\%} & \ccol{$\bm{\rho^2}$} & \ccol{Adj. $\bm{\rho^2}$} & \ccol{p-value} &  \ccol{power}\\
    \midrule
\multirow{3}{*}{UDPipe 1.2} & None & -0.492 & [-0.63 -0.32] & 0.242 & 0.234 & $<$0.001 & 0.999 \\
& Train Tokens & -0.265 & [-0.44 -0.06] & 0.070 & 0.050 & 0.010 & 0.735\\ 
& Train Tokens, \meanL & -0.278  & [-0.46 -0.08] & 0.077  & 0.047 & 0.007 & 0.773 \\
\midrule
\multirow{3}{*}{UDPipe 2.0} & None & -0.466 & [-0.61 -0.29] & 0.217 & 0.208 & $<$0.001 & 0.997 \\
&Train Tokens & -0.290 & [-0.47 -0.09] & 0.084 & 0.063 &  0.006 & 0.796 \\
&Train Tokens, \meanL & -0.312 & [-0.49 -0.11] & 0.097 & 0.066 & 0.003& 0.849\\
        \bottomrule
    \end{tabular}

\end{table}

We next use this technique to evaluate the relationship observed between EDV and LAS. In Figure \ref{fig:noise-signal} we show the fit of LAS against the number of training tokens (leftmost plot), and then the first normalized LAS against \meanL{} (middle plot), and the final normalized LAS against EDV (rightmost plot) for UDPipe 2.0 (Figure \ref{fig:noise-signal-p1} in \hyperref[appendix:vis]{Appendix~A} shows the equivalent analysis for UDPipe 1.2). We opted to include \meanL{} even though no correlation was observed between \meanL{} and EDV because theoretically it could impact the measurement of EDV, and if the coefficients failed to capture this, it could still impact the final analysis. However, removing the signal associated with it and the number of training tokens still results in a clear linear relationship between EDV and LAS (correlation of $-$0.283 with p-value$=$0.007). The correlation is much diminished compared to the original coefficient measured for EDV of $-$0.466 (Table \ref{tab:las_spearmans}), but it is still meaningful. The results are echoed in the analysis for UDPipe 1.2 with a correlation of $-$0.249 (p-value$=$0.015) between EDV and the final normalized LAS compared to $-$0.492 for the original measured coefficient (Table \ref{tab:las_spearmans}).

\subsubsection{Partial coefficients}\label{sec:partial}

This ultimately leads us to evaluating EDV with respect to LAS using partial coefficients. The main covariant of interest is the number of tokens in the training data, which is not only the most strongly correlated variable with respect to LAS (Table \ref{tab:las_spearmans}) but also the second most strongly correlated variable with respect to EDV (Table \ref{tab:inter_spearmans}). We also include \meanL{} despite measuring no correlation with it and EDV because of the apparent impact it had in the background subtraction analysis (Section \ref{sec:bcg}). In Table \ref{tab:re_results}, we show the full measurement of the partial coefficients for EDV with respect to LAS for UDPipe 1.2 and 2.0 with no covariants (i.e., the standard coefficient), with the number of training tokens as the sole covariant, and with both the training tokens and \meanL{} as covariants. As expected, when evaluating the correlation with the number of training tokens as a covariant we observe the biggest change in the measured coefficient. For UDPipe 1.2 it drops from $-$0.492 to $-$0.265 and for UDPipe 2.0 it drops from $-$0.466 to $-$0.290. We also note that despite not being correlated based on the calculated coefficients between \meanL{} and EDV, we still checked its impact. There is a small increase in the partial coefficients here signaling that \meanL{} is not a covariant of EDV with respect to LAS. This partial correlation coefficient results in an adjusted $\rho^2$ of 0.047 for UDPipe 1.2 and 0.066 for UDPipe 2.0, which gives a less biased indication of the proportion of explained variance associated with EDV (5\% for UDPipe 1.2 and 7\% for UDPipe 2.0). Only including the number training tokens as a covariant results in an adjusted $\rho^2$ of 0.050 for UDPipe 1.2 and 0.063 for UDPipe 2.0 (5\% and 6\% respectively). Therefore, in this setting, we can say that EDV is correlated with a non-trivial amount of the differences observed in parsing performance across treebanks.\footnote{However, interpreting correlations is somewhat subjective. Others might see these values and surmise that EDV is less informative than the training data size on its own and only adds a small amount of additional explanation of the observed variation in parsing performance. We have attempted to report the statistics in a way that readers can come to their own conclusions while also offering our personal interpretations.}

\begin{table}[t!]
    \centering
       \footnotesize
           \caption{Statistics associated with linear regression models using combinations of log size, EDV, and \meanL{} as predictors. We report the adjusted R$^2$ scores for linear regression fits as a less biased indication of the proportion of explained variance and report the percentage of relative importance of each predictor along with the corresponding p-values.} 
    \label{tab:linear_model}
       \tabcolsep=0.13cm
    \begin{tabular}{llccc}
        \toprule
\textbf{Parser} & \textbf{Variables} & \ccol{Adj. R$^2$} & \ccol{Relative Importance} & \ccol{p-values} \\\midrule
\multirow{4}{*}{UDPipe 1.2} & $\log$Train Tokens & 0.475 & 100.0 & $<$0.001\\
& $\log$Train Tokens, \meanL & 0.503 & 87.8, 12.2 & $<$0.001, 0.015\\
& $\log$Train Tokens, EDV & 0.567 & 55.7, 44.3 & $<$0.001, $<$0.001\\
& $\log$Train Tokens, \meanL, EDV & 0.589 & 50.8, 8.6, 40.6 & $<$0.001, 0.018, $<$0.001\\
\midrule
\multirow{4}{*}{UDPipe 2.0} & $\log$Train Tokens & 0.434 & 100.0 & $<$0.001\\
& $\log$Train Tokens, \meanL & 0.468 & 88.3, 11.7 & $<$0.001, 0.012\\
& $\log$Train Tokens, EDV & 0.494 & 61.4, 38.6 & $<$0.001, 0.001\\
& $\log$Train Tokens, \meanL, EDV & 0.522 & 56.0, 9.1, 35.0 & $<$0.001, 0.015, 0.001\\
\bottomrule
    \end{tabular}
\end{table}
\subsection{Multilinear regression}\label{sec:linear}

We then evaluated the impact EDV has in a multilinear regressive fit of the data for both systems. The results are shown in Table \ref{tab:linear_model}. We start by simply fitting a model using the log of the number of training tokens and for both systems we obtain a fit that has reasonably large adjusted R$^2$ (0.475 and 0.434 for UDPipe 1.2 and 2.0, respectively). We also use \meanL{} based on the results from Sections \ref{sec:bcg} and \ref{sec:partial} and see that the adjusted R$^2$ for the model using this and the log of training token size is slightly higher than only using the training tokens (about 0.03 for both systems). Using training tokens with EDV, however, results in a larger increase of 0.09 for UDPipe 1.2 and 0.06 for UDPipe 2.0. We also observe an increase when using EDV in addition to the other two variables which results in the largest adjusted R$^2$ of 0.589 and 0.522 for UDPipe 1.2 and 2.0, respectively. 

It is necessary to highlight that despite reporting the adjusted R$^2$, it is still a biased indication of the proportion of explained variance of a model. However, it is still indicative of the quality of the model, but more importantly it allows us to evaluate the impact of EDV. We also report the relative importance percentages in Table \ref{tab:linear_model} which show that EDV  roughly carries 40\% of the importance in the models it is used in for UDPipe 1.2 and about 35\% for UDPipe 2.0. 
\subsubsection{Testing the model with UDPipe 1.2}
\begin{figure}[t!]
    \centering
    \includegraphics[width=0.55\linewidth]{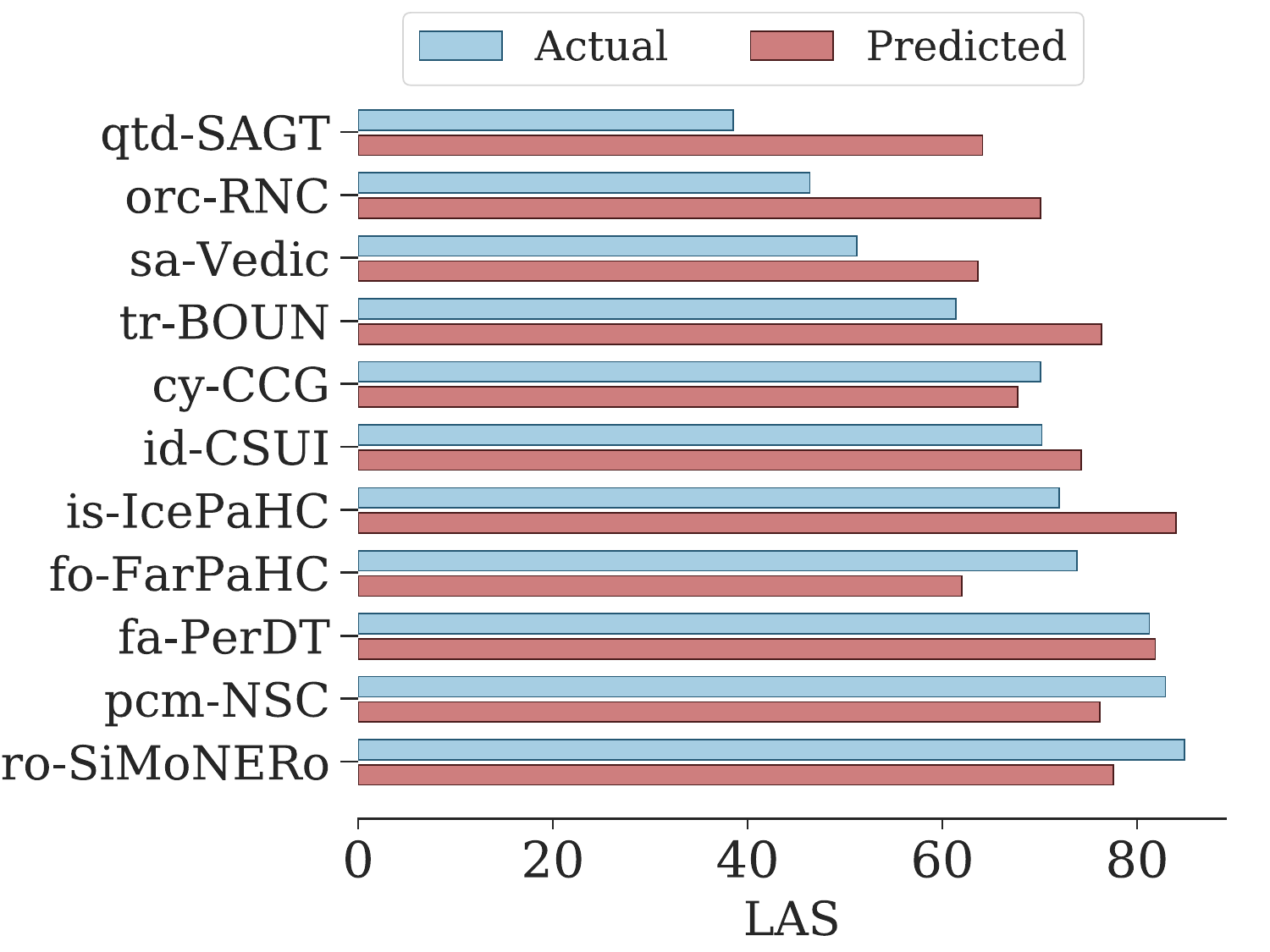}
    \caption{Comparison of performance of new UDPipe 1.2 models for treebanks not covered in current UDPipe 1.2 models that appear in UD v2.7 with predictions from linear model from Section \ref{sec:linear} using the log of the number of training tokens, \meanL, and EDV as predictors. The mean absolute error is 11.05.}
    \label{fig:predicted}
\end{figure}
As there exists a more up-to-date version of UD that contains more treebanks not used in the systems we have evaluated, we can use these new treebanks to evaluate the linear model from Section \ref{sec:linear}. We select the new treebanks based solely on two criteria: that the treebanks have at least 100 training sentences (as  very small treebanks tend to be very volatile with respect to performance) and that they contain pre-existing training and test sets (and potentially a development set). This resulted in 11 new treebanks. Note, Latin-LLCT fit these criteria but we opted not to use it as it contains the same sentence 356 times across the training,  development, and test data. 

We trained models using UDPipe 1.2 with the general settings. This means these data points are slightly different from those used to develop the linear regression model which were all optimized for each treebank based on the algorithm and oracle used. We ran the evaluation the same as described in Section \ref{sec:parsers}. We did not train models for UDPipe 2.0 as the parser is not publicly available. We then compared the LAS we obtained from these parsers and the values predicted by the linear regression model using all 3 variables as discussed in Section \ref{sec:linear}. The comparisons are shown in Figure \ref{fig:predicted} where the predicted values are not outlandishly different for most treebanks except for those which obtained fairly low LAS. While we have not set out to develop a predictive model, this is still useful as  a sanity check (if the predictions had been wildly inaccurate across the board, then one would have to question not only the linear model but the calculated coefficients).

\subsection{Sentence length binning}
\begin{figure}[t!]
    \centering
    \includegraphics[width=0.55\linewidth]{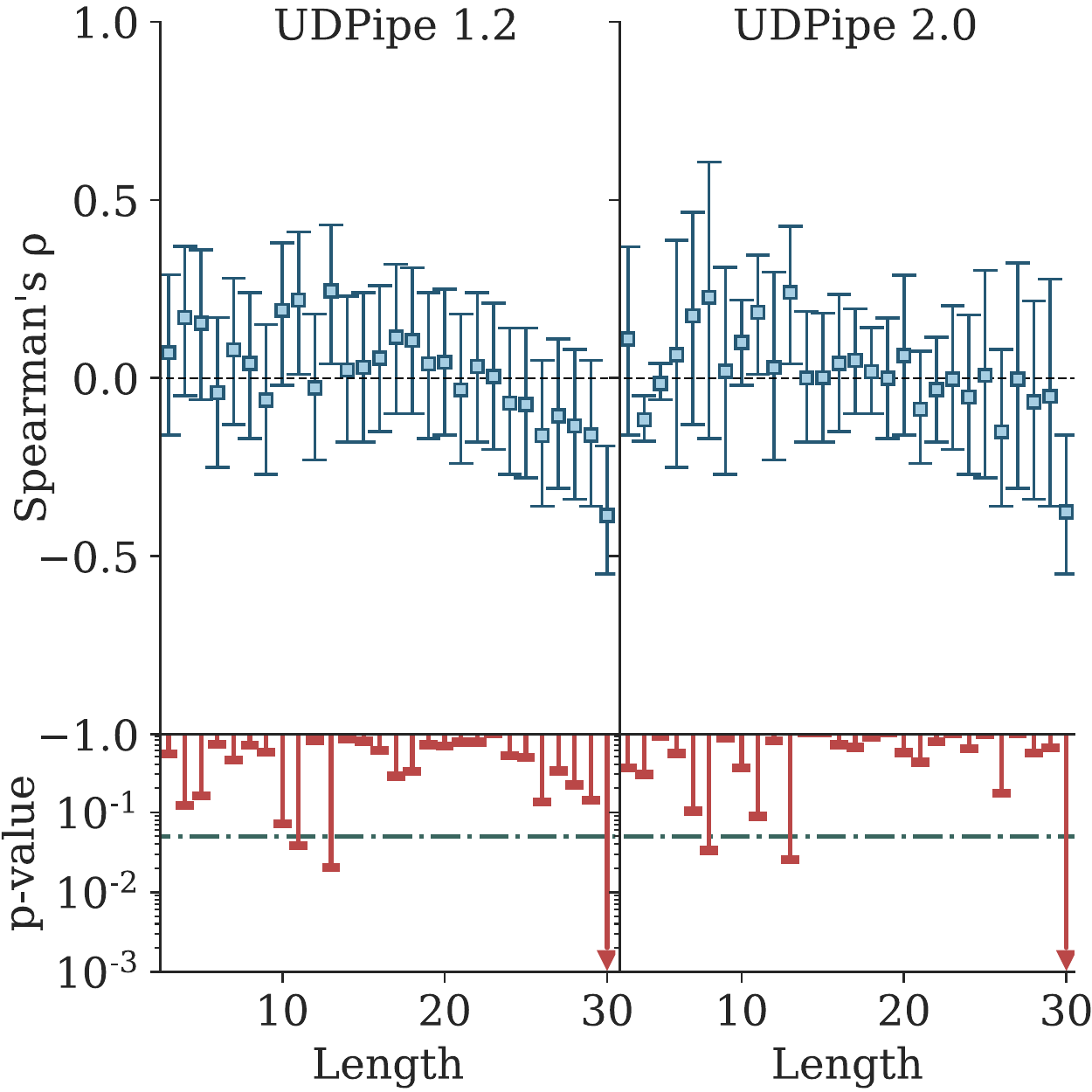}
    \caption{Partial correlation coefficients (top, blue) and their corresponding p-values (bottom, red) for UDPipe 1.2 (left) and UDPipe 2.0 for sub-samples binned with respect to sentence length. Comparison is between the EDV and LAS of each sub-sample with training tokens as covariant.}
    \label{fig:binned_partial_corrs}
\end{figure}

Here we turn to our sentence length binning analysis. As shown above in Figure \ref{fig:edda_vs_length} (and Figure \ref{fig:edda_vs_length_25} in \hyperref[appendix:vis]{Appendix~A}), EDV does show an expected dependency on sentence length. We also would like to highlight that this dependency is hardly unique to this situation, but consideration of this is almost completely lacking in NLP. Figure \ref{fig:binned_partial_corrs} shows the partial correlation coefficients and the corresponding p-values for each sentence length bin we evaluated in this analysis (sentence lengths of 3 to 30) for both parsers. Note that we only used the number of tokens in the training data as a covariant because for each bin \meanL{} is constant across each treebank by design. 

A clear trend can be observed where the magnitude of the correlations increases as a function of sentence length. However, most correlations don't have a particularly low p-value with the largest sentence-length bin being the exception. We offer visualization of the corresponding scatter plots for each bin in Figures \ref{fig:edda_binned_1} and \ref{fig:edda_binned_2} in \hyperref[appendix:vis]{Appendix~A} for UDPipe 1.2 and 2.0, respectively. From these plots, it appears that there are some linear relations that echo the correlation coefficients reported in Figure \ref{fig:binned_partial_corrs}, but these plots of course don't handle the number of training tokens. 
In this setting, EDV's interplay with parsing performance is not as convincing as in the other analyses. This could be related to issues with having less data as a necessary result of the binning procedure, therefore impacting the reliability of the statistics. It could also be that it introduces wider variances with respect to amount of training data in each sentence bin. It is clear that sentences of length 30 are correlated and have a coefficient that follows the trend, so it isn't as if EDV is completely uncorrelated in this setting. Whatever the reason for the different result observed here, this highlights the need to evaluate these exploratory correlation-based studies in different ways, so as to temper the certainty with which we present our results.

We note that unlike other treebank analyses focusing on measurements that are likely to be related to sentence length, EDV has a clear global correlation with our target variable (e.g., \citet{anderson-gomez-rodriguez-2020-inherent} did not observe a global correlation in their analyses). But the sentence length binning highlights that different signals can be found in a more fine-grained analysis.
\section{Morphological complexity}\label{sec:mc}

\begin{figure}[b!]
    \centering
    \includegraphics[width=0.8\linewidth]{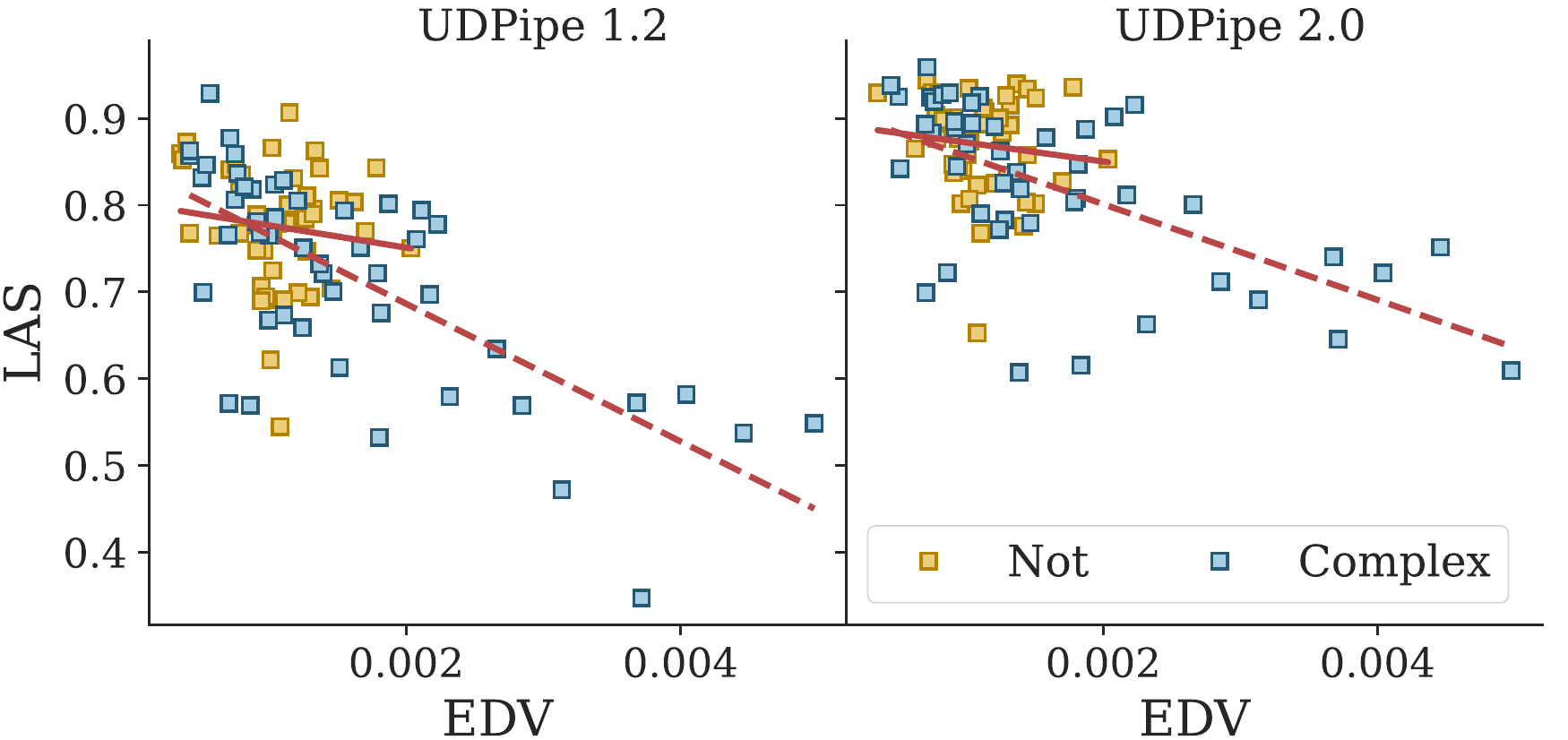}
    \caption{Visualization of LAS against EDV for UDPipe 1.2 and UDPipe 2.0 for the morphologically complex subset of treebanks (Complex, blue) and the not morphologically complex subset (Not, yellow). Linear fits are shown to aid visualization (continuous line for not complex and dashed for complex).}
    \label{fig:morph_results}
\end{figure}

Here we offer a small analysis of a subset of the data which are measured to be morphological complex. We use an aggregate measurement that is explained in detail in \hyperref[appendix:complexity]{Appendix~C} to measure the morphological complexity of the training data in a given treebank. This consists of 5 metrics which have been normalized and calibrated such that for each measurement 0 means no morphological complexity and 1 means maximum complexity. The average is then taken of these 5 metrics. They are based on word entropy \citep{shannon1948a}, type--token ratio \cite{bentz2016a}, form to lemma ratio, form to inflected lemma  ratio, and head part-of-speech entropy \citep{dehouck-denis-2018-framework}. They all measure slightly different aspects of morphological production, except head part-of-speech entropy which measures morphosyntactic complexity. Mathematical descriptions of these measurements are given in \hyperref[appendix:complexity]{Appendix~C}, detailing the original measurements and how they have been normalized so that they could be more readily combined. For more details on these measurements (including experiments evaluating the interplay between them and parsing) see \citet{bentz2016a}, \citet{dehouck-denis-2018-framework}, and \citet{Dehouck2019MultiLingualDP}.

We simply take the most morphologically complex treebanks by considering a treebank \textit{morphologically complex} if its complexity is greater than the mean measurement across treebanks. This results in 50 \textit{morphologically complex} treebanks in UD v2.5 (out of 94) and    47 in UD v2.6 (out of 90). Lists containing the specific treebanks considered morphologically complex are given in \hyperref[appendix:complexity]{Appendix~C}. We cut it this way as we did not find other reasonable arguments for applying a different threshold. They are all equally arbitrary. At least following this criterion we split in a way that doesn't introduce any biases (outside of the data). It does result in some treebanks from similar languages (or the same) appearing in different subsets, e.g. Portuguese-GSD has a result of 0.60 for the aggregate score, Portuguese-Bosque has 0.52, Galician-TreeGal has 0.55, and the mean score is 0.57 which results in Portuguese-GSD being classed as morphologically complex and Galician-TreeGal and Portuguese-Bosque as not. However, this measurement is not meant to classify languages but to compare given samples of a language that appear in treebanks.
Furthermore, if a given property (in our particular case, correlation between LAS and EDV) tends to hold for morphologically complex treebanks and not the others (or vice versa), the fact that a treebank of intermediate complexity falls on one or other side of the split should have little influence on the aggregate metrics that we use to detect this, as long as clear-cut cases are assigned to the correct subset.
\begin{table}[t!]
    \centering
        \footnotesize
        \caption{Partial coefficients (except for rows with None in covariant column) for the full set of treebanks (Full), the morphologically complex subset (Com.), and the not morphologically complex  subset (Not) for EDV with respect to LAS for UDPipe 1.2 and UD v2.5 (top) and for UDPipe 2.0 and UD v2.6 (bottom). Shown is the coefficient itself ($\rho$), the 95\% confidence interval (CI95\%), $\rho^2$ as an indication of the proportion of explained variance, the adjusted $\rho^2$ (Adj. $\rho^2$) as a less biased version of $\rho^2$, the corresponding p-values, and the achieved power of the test (power).}
    \label{tab:morph_results}
    \color{black}
    \begin{tabular}{lllrrcrrrr}
    \toprule
\textbf{Parser} & Set & N &\textbf{Covar.} &  \ccol{$\bm{\rho}$} &  \ccol{CI95\%} & \ccol{$\bm{\rho^2}$} & \ccol{Adj. $\bm{\rho^2}$} & \ccol{p-value} &  \ccol{power}\\
    \midrule
\multirow{6}{*}{UDPipe 1.2} &Com. & 50 & None & -0.678 & [-0.80 -0.49] & 0.459 & 0.448 &  $<$0.001 & 1.000 \\
& Not & 44 & None &  -0.073 & [-0.36 \enskip0.23] & 0.005 & -0.018  & 0.636 &  0.076 \\ 
 & Full & 94 &None & -0.492 & [-0.63 -0.32] & 0.242 & 0.234 & $<$0.001 & 0.999 \\\cmidrule{2-10}
&Com. & 50 &Train Toks & -0.481 & [-0.67 -0.23] & 0.231 & 0.199   & $<$0.001 & 0.948 \\ 
 & Not & 44 &Train Toks & 0.163 & [-0.14 \enskip0.44] & 0.027 & -0.021 & 0.296 & 0.182 \\
& Full & 94 &Train Toks & -0.265 & [-0.44, -0.06] & 0.070 & 0.050 &  0.010 & 0.735 \\
\midrule
\multirow{6}{*}{UDPipe 2.0} &Com. &  47 &None & -0.624 & [-0.77 -0.41] & 0.389 & 0.376 & $<$0.001 & 0.998\\
& Not & 43 &None & -0.118 & [-0.40 \enskip0.19] & 0.014 & -0.010 &  0.450 & 0.118 \\ 
 & Full & 90 &None & -0.466 & [-0.61 -0.29] & 0.217 & 0.208  & $<$0.001 & 0.997 \\\cmidrule{2-9}
&Com. & 47  &Train Toks & -0.466 & [-0.64 -0.16] & 0.183 & 0.146 & 0.003 & 0.857 \\ 
 & Not & 43  &Train Toks & -0.008 & [-0.31 \enskip0.30] & 0.000 & -0.050 & 0.958  & 0.050 \\
& Full &  90  &Train Toks & -0.290 & [-0.47 -0.09] & 0.084 & 0.063 & 0.006 & 0.796 \\ 
\bottomrule
    \end{tabular}
\end{table}

Table \ref{tab:morph_results} gives the correlation coefficients for the two subsets of the data, the morphologically complex and the not morphologically complex, along with those for the full data. It is very clear that the morphologically complex subset has the clearest association with parsing performance for both parsers with an adjusted $\rho^2$ of 0.448 for UDPipe 1.2 and 0.376 for UDPipe 2.0, whereas the not morphologically complex subset has a negative $\rho^2$ for both (signaling that there is no linear relation). This clear relation holds even when accounting for the size of the training data with an adjusted $\rho^2$ of 0.199 for UDPipe 1.2 and 0.146 for UDPipe 2.0. It is clear from the visualization in Figure \ref{fig:morph_results} that this is due to the morphologically complex subset having a wider range of EDV values with many having much higher EDV values than the small values exclusively observed from the not morphologically complex subset. It is also very clear from this visualization that it is not necessarily the case that a morphologically complex treebank will exhibit large discrepancies between samples with respect to the edge displacement distributions, i.e., many treebanks in the morphologically complex subset have very small EDV values.

\section{EDV for evaluation}
Having established that EDV does correlate to parsing performance when accounting for covariants in a number of ways, we turn to a proof of concept for a potential application of EDV in NLP: using it to inform a more linguistically motivated means of creating \textit{adversarial} splits. We note here that large EDV values between samples for a given language likely capture a linguistic feature of that language, in that large samples that deviate to a great extent suggest that language is more syntactically volatile than others. This could be true across the board or it could be a matter of greater variety in syntactic structures in a given domain. 
However, differences can also occur intra-domain based on author preferences. 

We mention this here because recent work on developing adversarial splits focused on sentence lengths \cite{soegaard2020we}. This was an extension from criticism based on using standard splits, where random splits were suggested instead \cite{gorman-bedrick-2019-need}. Together these analyses showed that standard splits and random splits are not enough to truly evaluate the brittle nature of NLP systems trained on data from a narrow set of domains. \citet{soegaard2020we} found that even when evaluating systems with adversarial splits (based on sentence length), the evaluation over-estimated the performance of the systems when compared with fresh samples.
We argue that creating adversarial splits based on sentence length is only weakly linguistically motivated (i.e., the variance in sentence length could be associated with different domains, but maximizing the difference between test and training set is only a very coarse approximation of differences in domain, as not all long sentences are necessarily harder for a model to handle). 
With this in mind, we propose using EDV to guide sampling to create adversarial and complementary splits to give an approximation of the volatility of parsing performance. As highlighted by \citet{soegaard2020we}, this only offers us a clearer picture of the generalizability of models based on the data available, which often overestimates the quality of models. However, in lieu of fresh data, this offers us a clear path to a more robust evaluation. 

This approach is not dependent on the parser being evaluated as the parser does not directly play a role in developing the splits. Despite this, it is clear that splitting on EDV might not offer robust evaluation for \textit{all} parsers and \textit{all} parser types, e.g, parsers that are not data-driven might be less sensitive to EDV. But if that were to be the case (certain parsers being less sensitive to differences in EDV), it could be argued that such parsers \textit{are} more robust than others. Finally, although we suggest this sampling method for evaluating parsers (and potentially other NLP systems), we are not suggesting this sampling method be the \textit{only} means of evaluating the generalizability of models. 
\subsection{Sampling}
We sample data in such a way so as to minimize EDV and maximize EDV, in order to give certain empirical limits of performance for a given treebank. We do this by collating all trees for a given treebank across all splits that are available. We remove trees with 2 tokens or less. We then bin the trees by sentence length and by the mean edge displacement (MED) of each sentence. MED is defined as: 
\begin{equation}\label{eqn:disp}
     \textrm{MED} = \frac{1}{N-1}\sum_{n\in N} s_{edge}^n
\end{equation}
where $n$ is a given node in a given tree, $N$ is the total number of nodes in the tree, and $s_{edge}^n$ is the edge displacement of a given node as defined in Equation \ref{eqn:displacement}. Note the denominator is $N-1$ as the root node is not included.

\begin{figure}[t!]
    \centering
    \includegraphics[width=0.49\linewidth]{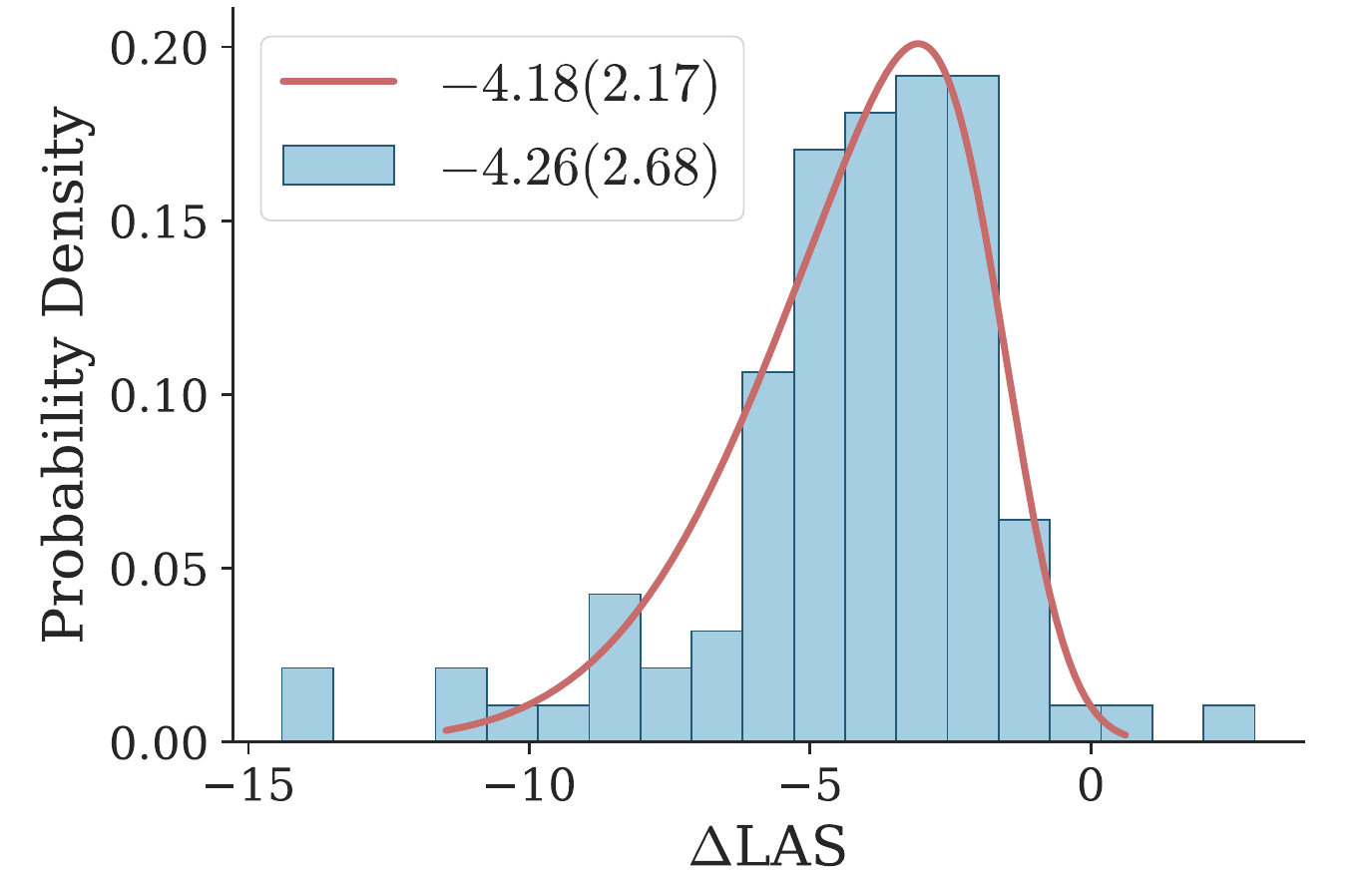}
    \caption{Distribution of  $\Delta$LAS (the LAS obtained from split where EDV is minimized minus the LAS obtained for the split where EDV is maximized) for UDPipe 1.2 models trained using UD v2.7 (103 treebanks). Shown is a fit used to obtain a more conservative measure of the variance between splits with $\chi^2=0.40$ and p-value$=$0.820 (note $H_0$ is that the data comes from the distribution described by the fit).}
    \label{fig:delta_las}
\end{figure}

We initialize the process by selecting a sentence length at random and also an MED value that exists for that sentence length bin. We then sample 3 more sentences with the closest sentence length and closest MED value available. This gives us 4 sentences with the same (or similar) sentence length and the same (or similar) MED. These are added to the training trees. We then either sample a sentence to match the MED value (when trying to keep EDV low) or sample a sentence with the furthest MED value available for the current sentence length bin (or closest if no sentences are left in a given bin) in order to maximize EDV. We repeat this process with the subsequent MED values chosen for the training instances to match the overall MED of the current training data. We do this until we have split the whole data into 80\% training data and 20\% test data. We then split the training data so as to obtain development data such that the overall split is 60$|$20$|$20 for training, dev, and test data, respectively. Note, we use MED and sample by tree as a more direct use of EDV would require the creation of many samples and hoping that one serendipitously maximizes/minimizes EDV. One could also potentially use an evolutionary algorithm to find splits which maximize (or minimize) EDV, but it would likely be computationally expensive. 

\begin{figure}[b!]
    \centering
    \includegraphics[width=0.45\linewidth]{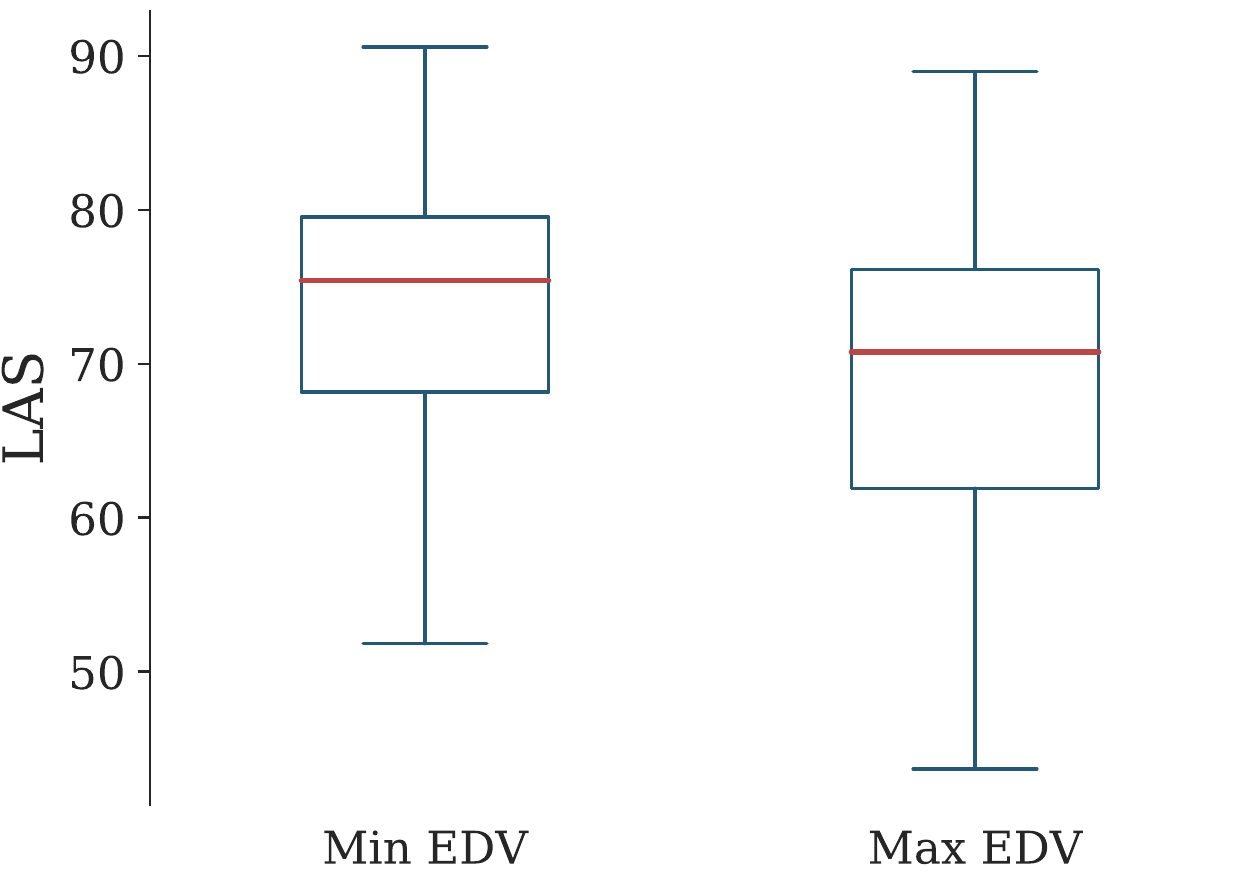}
    \caption{Distribution of LAS for UDPipe 1.2 models trained on splits sampled so as to miminize EDV (Min EDV) and sampling so as to maximize EDV (Max EDV) using UD v2.7 (103 treebanks). The median LAS for Min EDV is 75.40 and 70.77 for Max EDV.}
    \label{fig:sampled_boxes}
\end{figure}

We then train models using UDPipe 1.2 for the minimized EDV split and the maximized EDV split. We do this for all treebanks that have a training set of 100 sentences or more in the original split. 
\subsection{Sampling results}
Figure \ref{fig:delta_las} shows the distributions of $\Delta$LAS (LAS$_{max}-$LAS$_{min}$) for each treebank. We fit the distribution with a skewed Gaussian function to better evaluate variance seen in this process (a more conservative one at least). When evaluating the mean of the data itself we see a mean $\Delta$LAS of $-$4.26 (2.17) whereas the fit is slightly lower and with a higher standard deviation at $-$4.18 (2.68). This difference is considerable with typical claims of state-of-the-art performance coming down to tenths of a LAS point, so this process certainly gives a good range of performance across treebanks. Figure \ref{fig:sampled_boxes} shows the actual distribution of LAS values for both sets of splits. The median values of LAS are 75.40 and 70.77 for the minimum and maximum EDV splits, respectively. Note too that the spread across the first and second quartile is wider for the maximum EDV splits and with the lower tail being much smaller than that of the minimum split.

\begin{figure}[t!]
    \centering
    \includegraphics[width=0.515\linewidth]{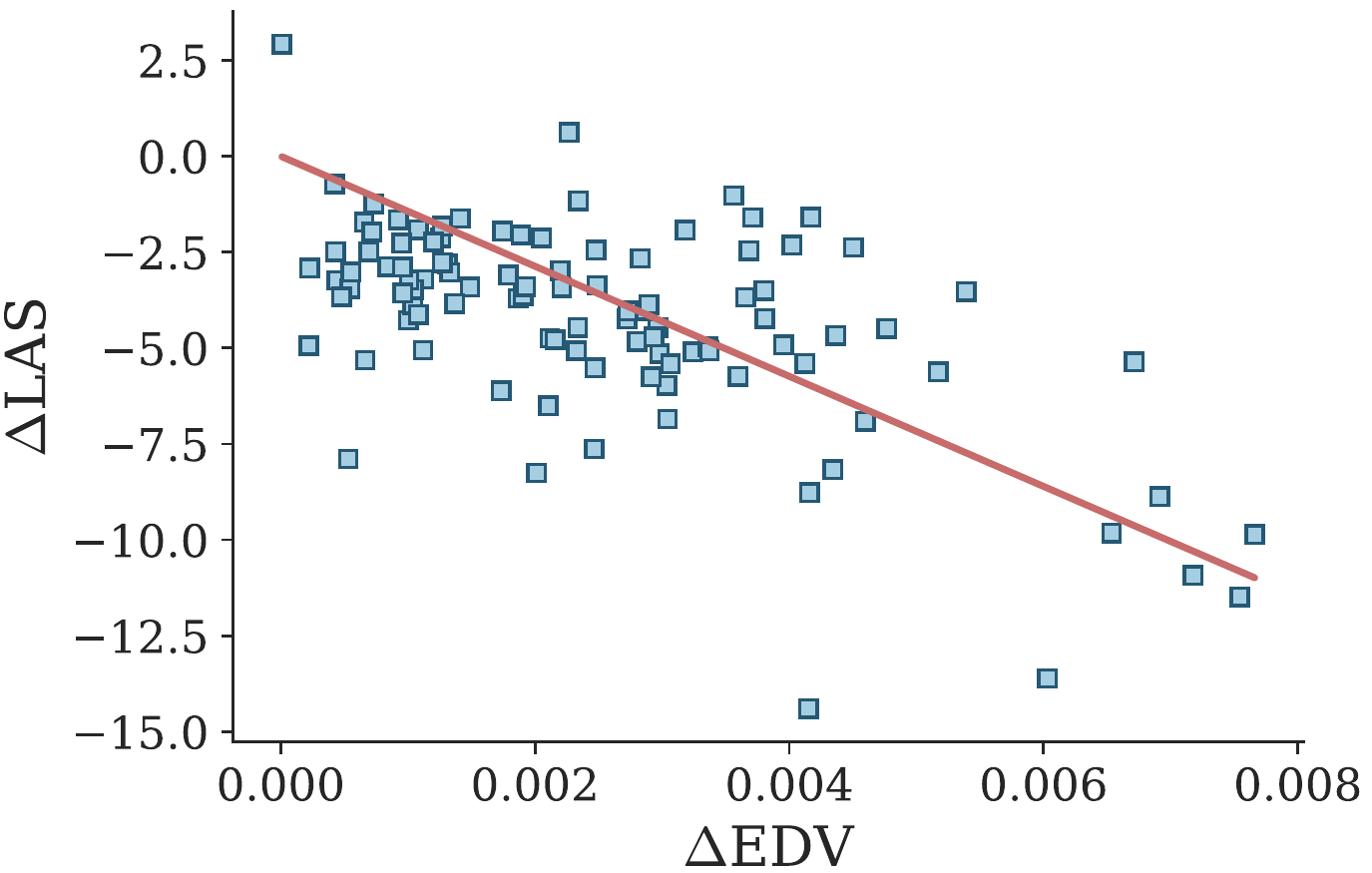}
    \caption{$\Delta$LAS against $\Delta$EDV where both are the value associated with the split where EDV has been minimized minus that of the split where EDV has been maximized. For UDPipe 1.2 models using UD v2.7 (103 treebanks).}
    \label{fig:delta_vs_delta}
\end{figure}

\begin{table}[b!]
\footnotesize
    \centering
        \caption{Correlations between variables of interest with respect to $\Delta$LAS using UD v2.7 (103 treebanks) and UDPipe 1.2. Shown are the coefficients ($\rho$), the 95\% confidence intervals (CI95\%), $\rho^2$ as an indication of the proportion of explained variance, the adjusted $\rho^2$ (Adj. $\rho^2$) as a less biased version of $\rho^2$, the corresponding p-values, and the achieved power of the tests (power). The mean absolute $\Delta$Tokens is 45.0 (68.7) which is a relative difference 0.097\% (with respect to the split where EDV is minimized). Mean absolute $\Delta$\meanL is 0.059 (0.172) which is a relative difference of 0.25\% wrt. Tokens and \meanL{} used here are the average across both splits.}
    \label{tab:delta_corrs}
    \begin{tabular}{lccrcrrrr}
    \toprule
     \multicolumn{1}{l}{\textbf{Variable}} & \textbf{Target} & \textbf{Covar.} &  \ccol{$\bm{\rho}$} &  \ccol{CI95\%} & \ccol{$\bm{\rho^2}$} & \ccol{Adj. $\bm{\rho^2}$} & \ccol{p-value} &  \ccol{power}\\
    \midrule
    Train Tokens & \multirow{6}{*}{$\Delta$LAS} & \multirow{6}{*}{---} & 0.104 & [-0.09, 0.29] & 0.011 & 0.001 & 0.295 & 0.182\\
    \meanL &  & & 0.507 & [0.35, 0.64] & 0.258 &  0.251 & $<$0.001 & 1.000 \\
    $\Delta$Tokens &  & & 0.067 & [-0.13, 0.26] & 0.004 & -0.006& 0.502 & 0.103 \\
    $\Delta$\meanL &  & & -0.037 & [-0.23, 0.16] & 0.001 & -0.009 & 0.713 & 0.065 \\
    $\Delta$SLV &  & & 0.139 &  [-0.06 \enskip0.32] & 0.019 & 0.010 & 0.161 & 0.290 \\
    $\Delta$EDV &  & & -0.478 & [-0.61, -0.31] & 0.228 & 0.220 & $<$0.001 & 0.999\\ \midrule
    \meanL & $\Delta$EDV & \multirow{2}{*}{---} & -0.847 & [-0.89, -0.78] & 0.717 &  0.711 & $<$0.001 & 1.000 \\
    $\Delta$SLV & $\Delta$EDV & & 0.088 & [-0.11 \enskip0.28] & 0.008 & -0.002 & 0.379 & 0.143\\
        $\Delta$EDV & $\Delta$LAS & \meanL & -0.105 & [-0.29 \enskip0.09] & 0.011 & -0.009 & 0.295 & 0.182\\
    $\Delta$EDV & $\Delta$LAS & $\Delta$SLV & 
    -0.497 &  [-0.63 -0.33] & 0.247 & 0.231 & $<$0.001 & 1.000 \\
    \bottomrule
    \end{tabular}
\end{table}

We also evaluate whether this difference in performance can be attributed to the differences in EDV between the splits. Figure \ref{fig:delta_vs_delta} shows $\Delta$LAS against $\Delta$EDV (EDV$_{max}-$EDV$_{min}$). A strong negative linear relationship is observed as expected. To validate this observation, we once again turn to correlation coefficients. These are reported in Table \ref{tab:delta_corrs}. We look at the variables deemed most pertinent to evaluate EDV from the preceding analysis in Section \ref{sec:corrs}. In this context, the number of training tokens (here we take the mean across the splits as an approximation) is not associated with the difference in performance across splits, which would only likely be the case if this had a major role in constraining the maximization of EDV. Similarly the difference between the number of training tokens is not correlated to $\Delta$LAS (the difference between splits is not large at a mean relative difference of 0.097\%).

However, \meanL{} (defined as the mean across splits) is strongly correlated to $\Delta$LAS ($\rho$=0.507, p-value$<$0.001) and even more so to $\Delta$EDV ($\rho$=0.847, p-value$<$0.001). This is likely due to the dependence on sentence length to vary EDV (see Figure \ref{fig:edda_vs_length} and Figure \ref{fig:edda_vs_length_25} in \hyperref[appendix:vis]{Appendix~A}). However, the difference between \meanL{} for each split is not correlated to  $\Delta$LAS, meaning that the difference observed is not merely due to the sampling procedure being forced to sample sentences of different length so as to maximize EDV. This is further attested to by the small mean relative difference between the splits of 0.25\%.

$\Delta$EDV is also strongly correlated to $\Delta$LAS at $-$0.478 (p-value$<$0.001) which fits with the trend observed in Figure \ref{fig:delta_vs_delta}. We also report the partial coefficient of $\Delta$EDV with respect to $\Delta$LAS with \meanL{} as a covariant. This results in a coefficient of $-$0.105 (p-value$=$0.295) which clearly shows the variation in EDV between the splits is strongly bounded by the sentence lengths of the data. In a sense, the sentence length distribution or \meanL{} dictates how much we can optimize the difference between the max and min EDV splits, but sampling splits so as to maintain a similar sentence length distribution across splits but sampling randomly for each sentence length bin is likely to result in easier splits than maximizing EDV. We also check to see if the difference between the sentence length distributions ($\Delta$SLV) diminishes the correlation between $\Delta$EDV and $\Delta$LAS if used as a covariant, but it doesn't (it is in fact slightly larger but this increase is meaningless). 

\section{Conclusion}
We have offered an analysis which has shown a clear correlation between the differences in the edge displacement distributions of training and test data in UD treebanks (as measured by the Vaserstein distance) and parsing performance (as measured by the labeled attachment score) by using a number of methods to falsify this hypothesis. We attempted to remove signals associated with covariants which were also correlated with LAS, but still observed a linear relationship between EDV and a normalized LAS. We use statistical methods to first evaluate the partial correlations of EDV and LAS when accounting for covariants and still observed meaningful coefficients. We also used multilinear regression to evaluate whether EDV adds any predictive power to models using these same covariants and measured small but meaningful contributions from EDV. In addition, we evaluated this linear model by training new parsers with one of the systems under investigation here on treebanks in the most recent release of UD which did not already have a model and obtained predictions that were not outlandish, especially for higher performing treebanks. Further, we evaluated the partial coefficients for EDV when using a sentence-length binning analysis and observed stronger coefficients for sentences of moderate length with a clear monotonic relationship between the magnitude of the correlation of EDV to LAS and sentence length. However, the p-values are fairly high with only the largest sentences (of 30 tokens) exhibiting a large and clear correlation to parsing performance. 

As mentioned above, we suspect EDV is indicative of parsing performance because it captures syntactic differences at the sample level which could be down to a number of reasons, spanning different syntactic structures being adopted in different domains to linguistic features of a language causing greater degrees of freedom in the tree structures found in different samples. Beyond linguistic considerations, the difference in performance observed due to EDV is likely to be explained by supervised techniques struggling to predict unobserved patterns as larger EDV values indicate differences in the tree patterns found in the training and test data.

Finally, we have shown the potential for using EDV to create splits to evaluate an advantageous and a disadvantageous (based on the available data) scenario that is likely to be more indicative of real-world usage of parsers where out-of-domain, unseen syntactic structures likely occur in the outer regions of the distributions seen in narrow training data sets. We envisage this analysis also being useful for other practices in NLP. For example it could be used for evaluating the difficulty of a given instance for curriculum learning for training parsers or for other NLP tasks,  that is batches measured for EDV based on the overall distribution in the training data. 

\appendix
\renewcommand\thefigure{A.\arabic{figure}} 
\renewcommand{\theHfigure}{A.\arabic{figure}}

\begin{figure}
    \centering
    \includegraphics[width=\linewidth]{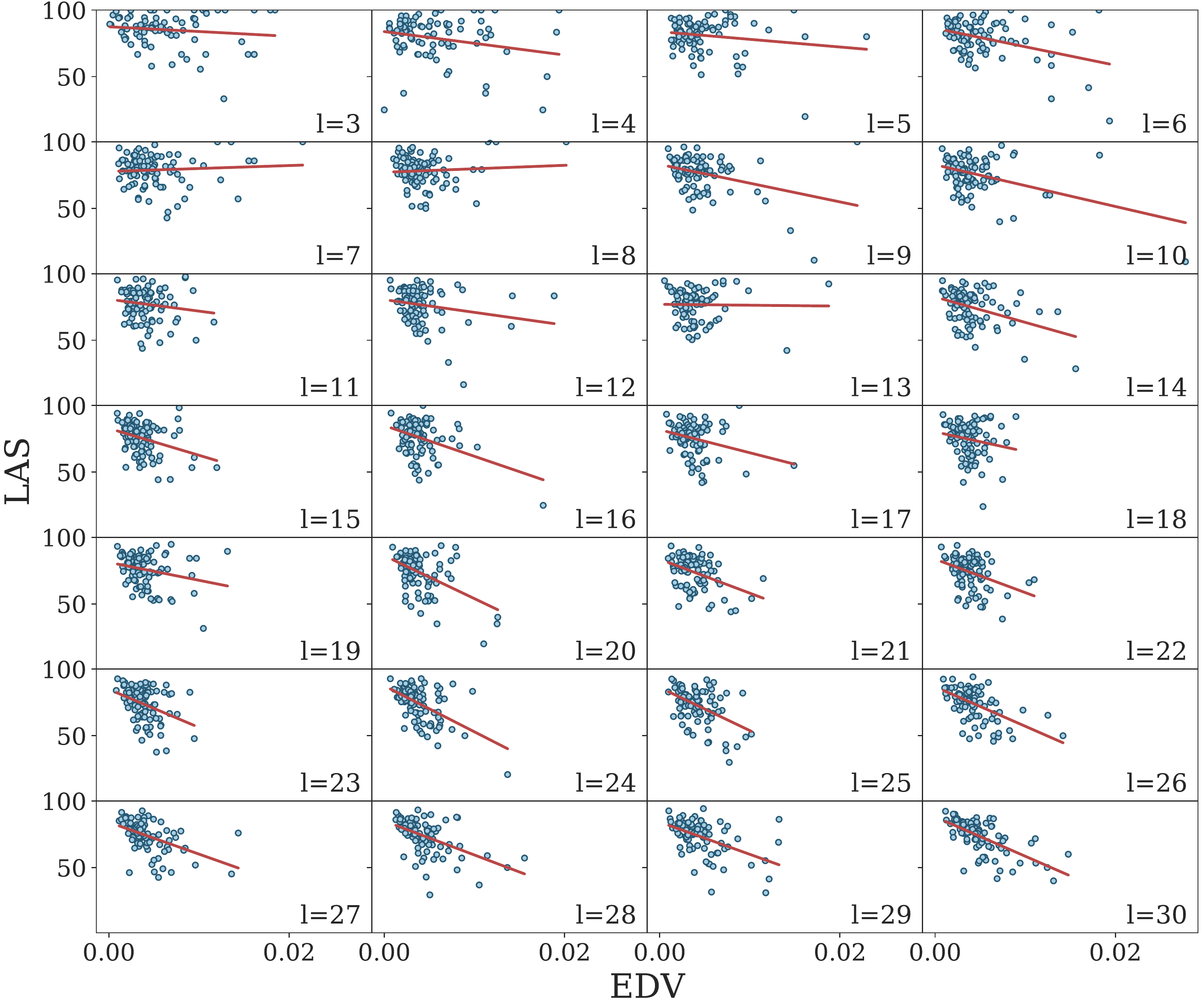}
    \caption{LAS versus EDV for each sentence length bin (labeled l=length) for UDPipe 1.2 used for calculating the coefficients shown in Figure \ref{fig:binned_partial_corrs}.}
    \label{fig:edda_binned_1}
\end{figure}

\begin{figure}
    \centering
    \includegraphics[width=\linewidth]{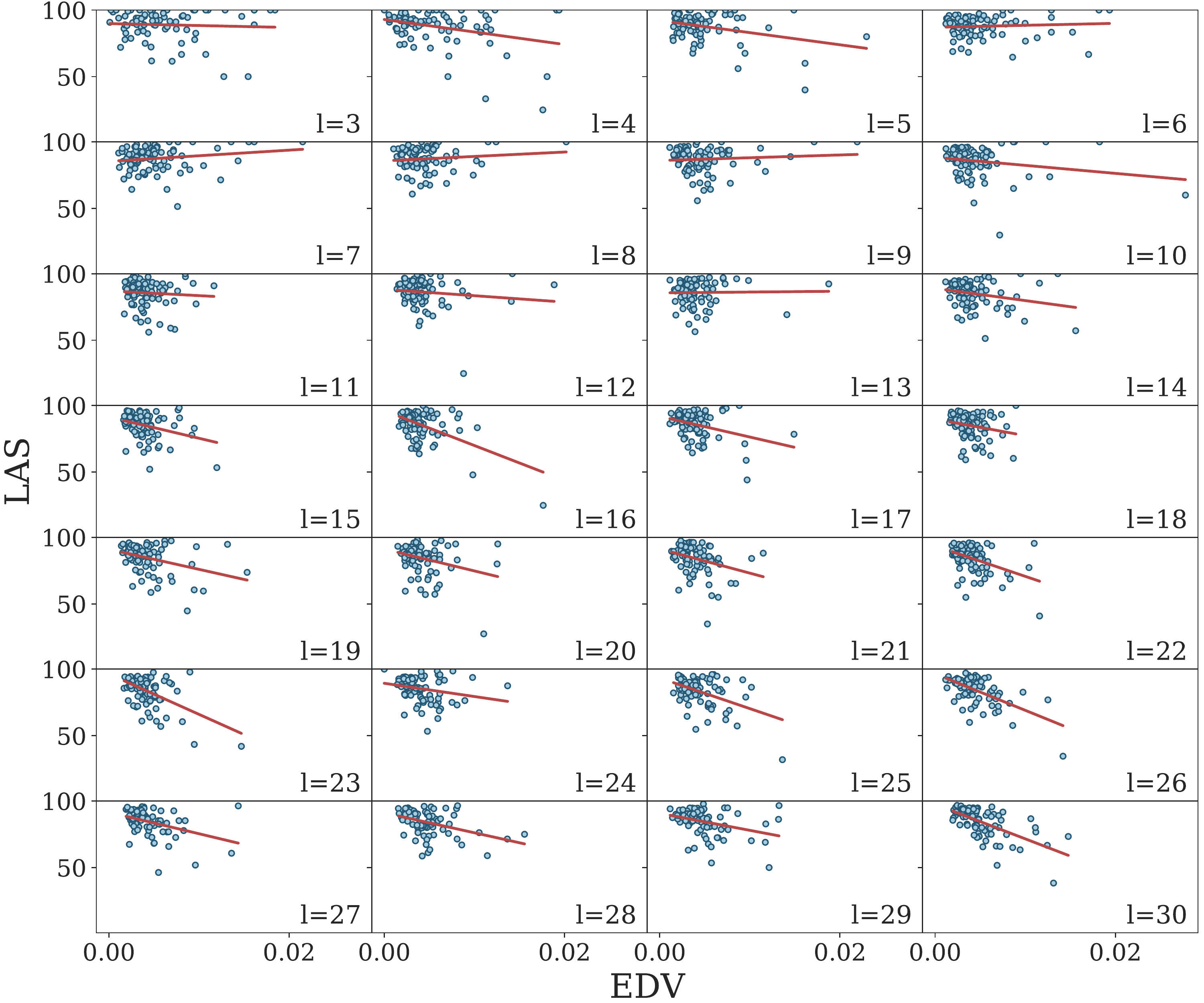}
    \caption{LAS versus EDV for each sentence length bin (labeled l=length) for UDPipe 2.0 used for calculating the coefficients shown in Figure \ref{fig:binned_partial_corrs}.}
    \label{fig:edda_binned_2}
\end{figure}

\appendixsection{Further visualization}\label{appendix:vis}
This appendix is mainly for showing the corresponding data for UDPipe 1.2 (as we showed the data for UDPipe 2.0 for the most part in the main text). Almost universally the observed behavior follows that shown in the main text. If it had been otherwise, we would have opted to show conflicting data visualizations. Figures \ref{fig:edda_binned_1} and \ref{fig:edda_binned_2} show the data used to evaluate the coefficients shown in Figure \ref{fig:binned_partial_corrs}.
  
\begin{figure}[htpb!]
    \centering
    \includegraphics[width=0.49\linewidth]{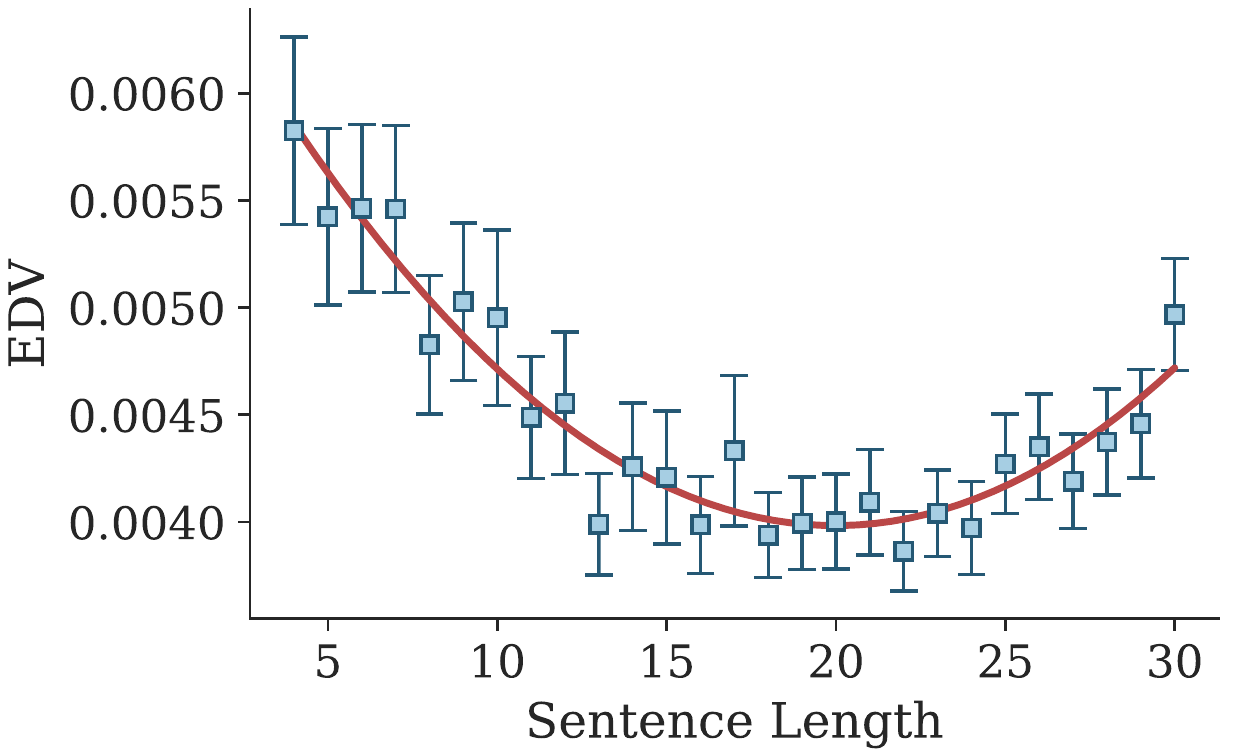}
    \caption{EDV between sub-samples of the training and test data binned by sentence length for UD v2.5 (105 treebanks).}
    \label{fig:edda_vs_length_25}
\end{figure}

    \begin{figure}[htbp!]
    \centering
    \includegraphics[width=0.49\linewidth]{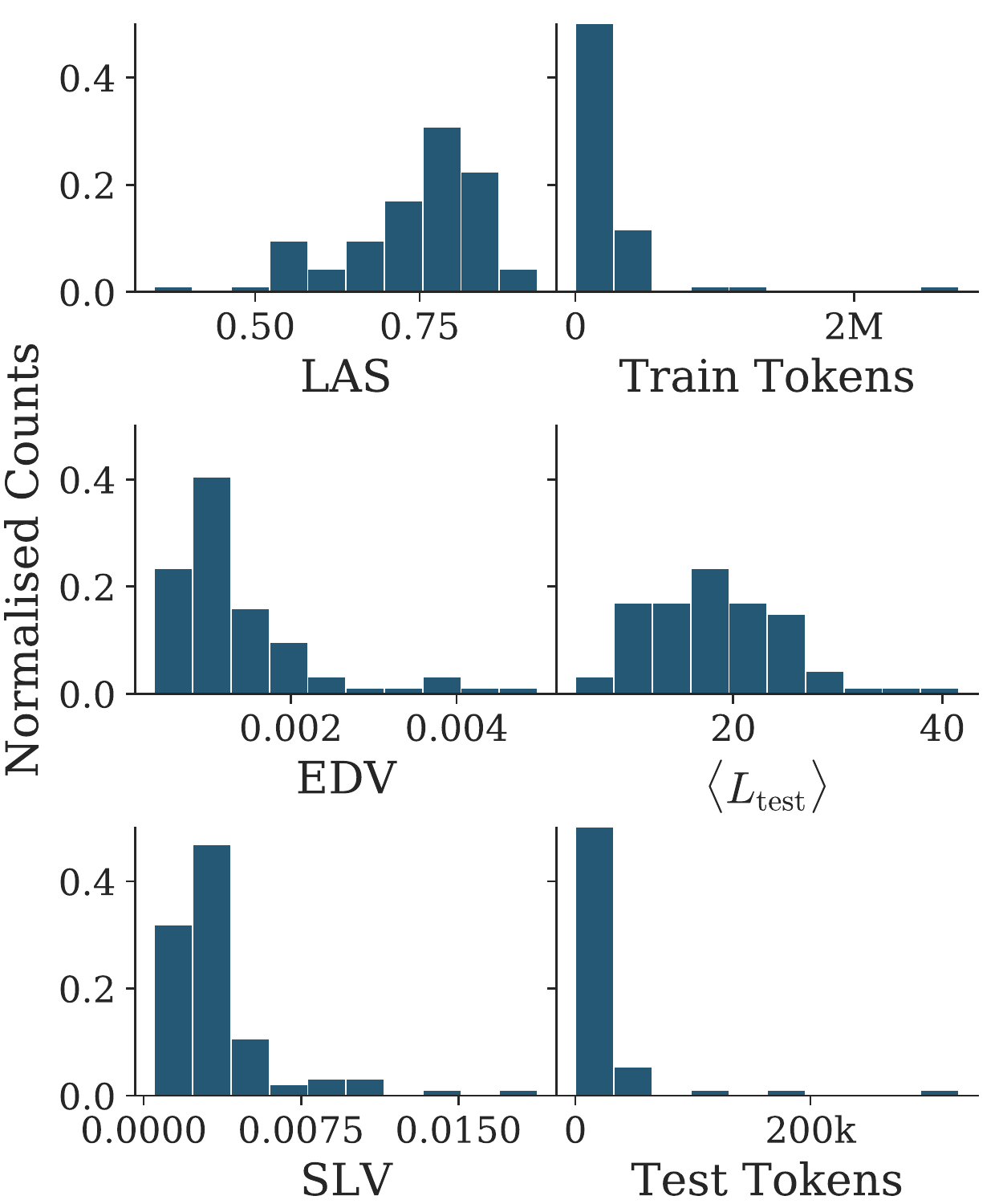}
    \caption{Distributions of the variables of interest in UD v2.5 (94 treebanks) in order to evaluate whether they are sampled from normal distributions.}
    \label{fig:not_normal_25}
\end{figure}
  
  \begin{figure}[htpb!]
    \centering
    \includegraphics[width=0.8\linewidth]{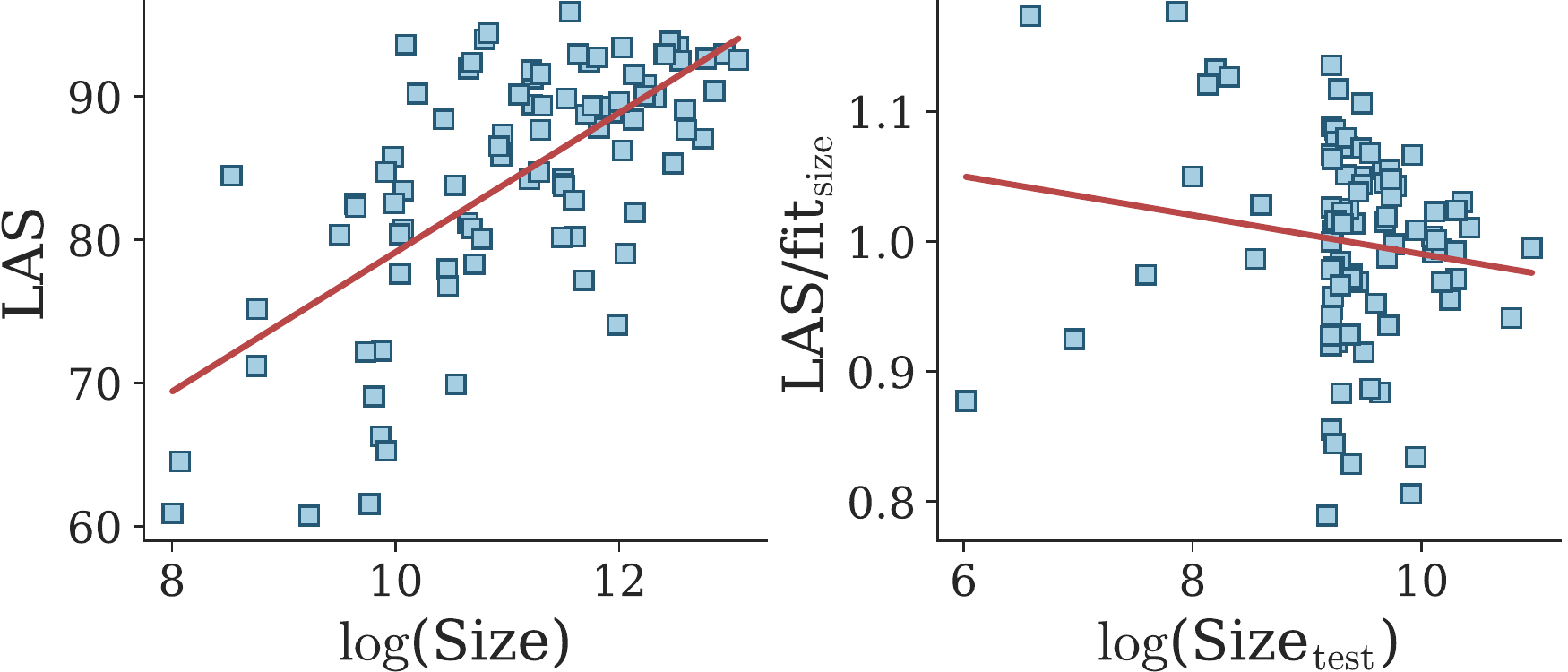}
    \caption{Background removing method used to evaluate whether the number of test tokens carries additional information with respect to the number of training tokens for UDPipe 2.0 and UD v2.6. Correlation between the number of test tokens and LAS is 0.309 (p-value$=$0.003) and that between the number of test tokens and the normalized LAS (right plot) is -0.101 (p-value$=$0.342).}
    \label{fig:bcg_test}
\end{figure}

\begin{figure}[htpb!]
    \centering
    \includegraphics[width=\linewidth]{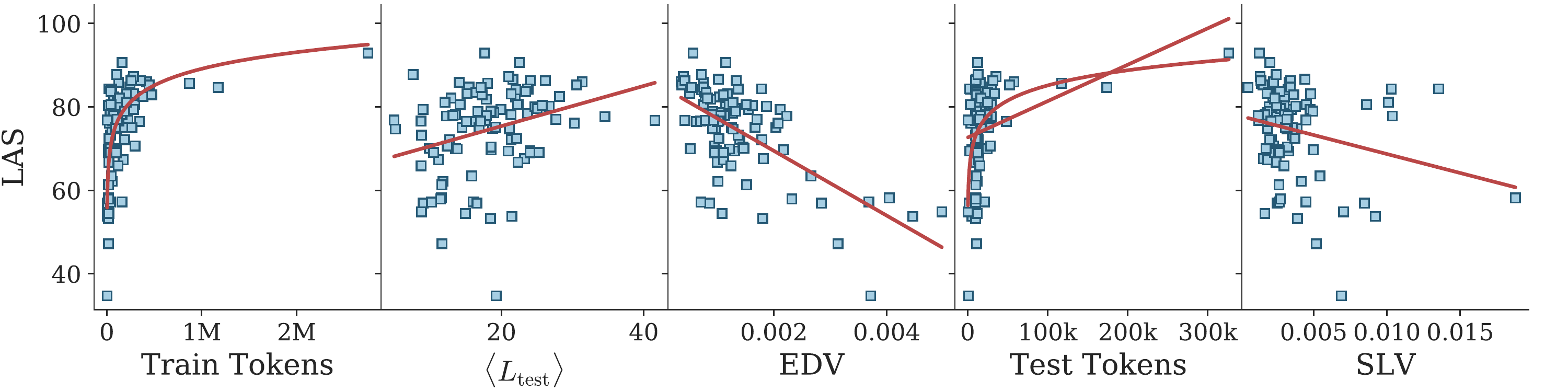}
    \caption{Visualization of LAS (for UDPipe 1.2 and UD v2.5) with respect to variables of interest with fits shown in red to highlight whether the data appears correlated or not.}
    \label{fig:correlations_wrt_las_p1}
\end{figure}

\begin{figure}[!thpb]
    \centering
    \includegraphics[width=0.82\linewidth]{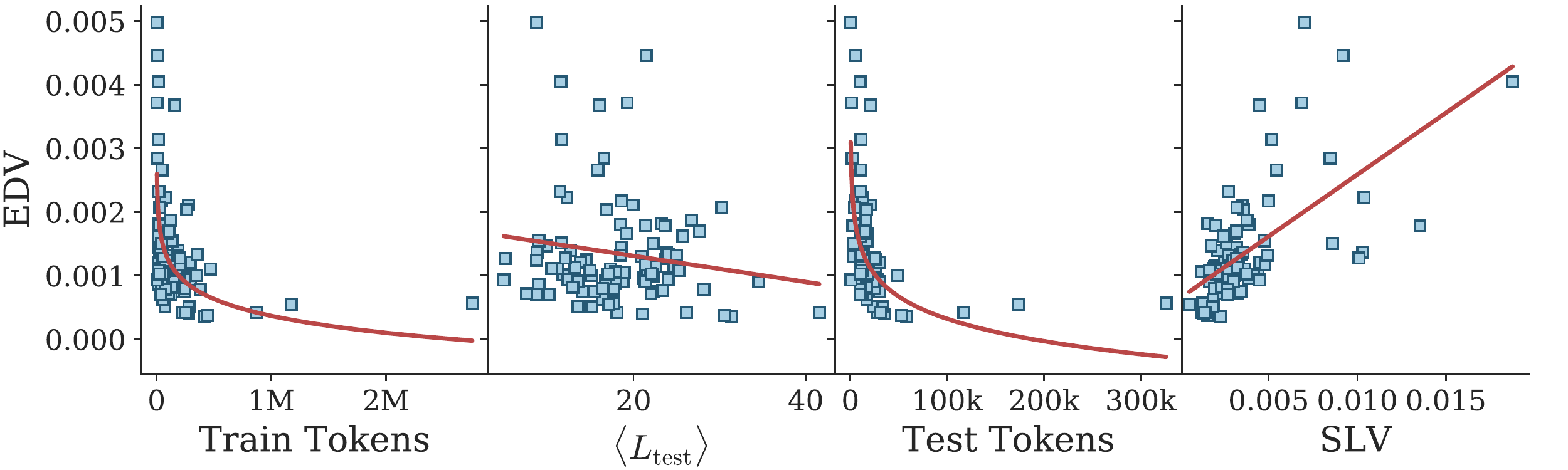}
    \caption{Visualization of EDV (for UD v2.5) with respect to variables of interest with fits shown in red to highlight whether the data appears correlated or not.}
    \label{fig:edv_vs_others_25}
\end{figure}

\begin{figure}[!thbp]
    \centering
    \includegraphics[width=\linewidth]{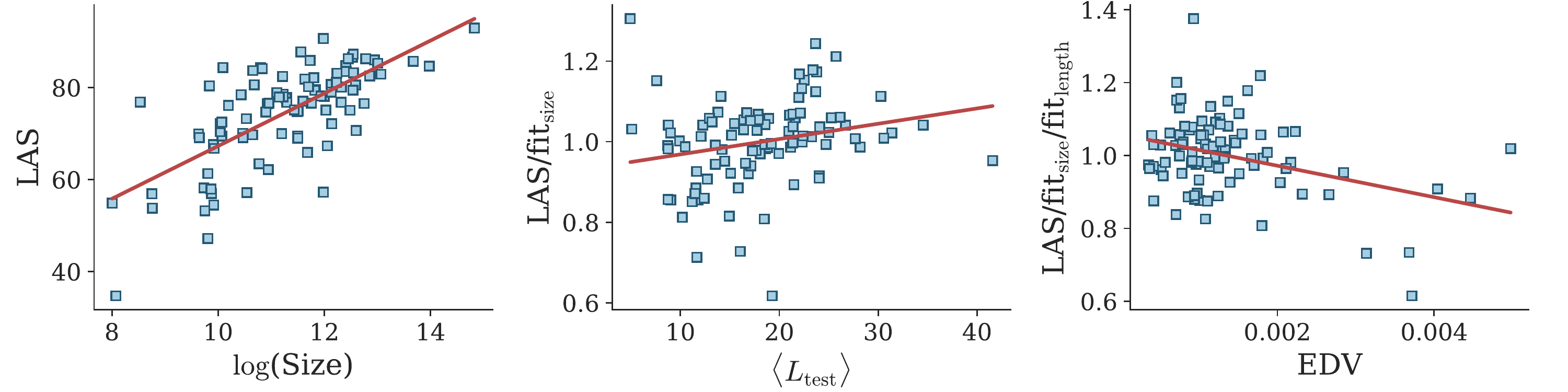}
    \caption{Background removal method to evaluate whether a correlation is observed between EDV and LAS (for UDPipe 1.2 and UD v2.5) after removing the variation associated with the training test size and \meanL. The correlation between EDV and LAS is -0.492 (p-value$<$0.001), the correlation between EDV and the LAS normalized by the variance associated with number of tokens in training data is -0.186 (p-value$=$0.072), and the correlation for the fully normalized LAS (removing the variance associated with \meanL) is -0.249 (p-value$=$0.015).}
    \label{fig:noise-signal-p1}
\end{figure}
\clearpage

\appendixsection{Training data measurements unrelated to EDV}\label{appendix:unrelated}
In this appendix, we include some additional analysis looking at some linguistically focused measurements of the training data which are related to parsing performance. These are presented here rather than the main text because there is no theoretical justification for expecting these measurements to impact the EDV of a given treebank split. The first metric is a normalized count of the number of crossings in a tree $C/|Q|$ where $C$ is the number of crossings in a tree and $|Q|$ is the total number of possible crossings \cite{ferrer2018crossing}. The second measurement is the type--token ratio defined as the number of unique forms divided by the number of tokens found in a treebank which gives a measure of the lexical diversity of a sample and a coarse indication of the degree of morphology. As can be seen in Table \ref{tab:ttr_crossings}, neither of these measurements are correlated to EDV for either UD v2.5 or UD v2.6. A visualization of the corresponding data is shown in Figure \ref{fig:reviewer_b}.
\renewcommand\thefigure{B.\arabic{figure}} 
\renewcommand\thetable{B.\arabic{table}} 
\begin{table}[htp!]
    \centering
    \small
     \caption{Spearman's $\rho$ for correlations between treebank measurements of training data and EDV. $C/|Q|$ is the normalized number of crossings found in a treebank and TTR is the type--token ratio.}
    \label{tab:ttr_crossings}
    \begin{tabular}{llrrrr}
    \toprule
   \textbf{Data} &\textbf{Variable}  & \ccol{$\rho$} & \ccol{CI95\%} & \ccol{p-value} & \ccol{power} \\ \midrule
   \multirow{2}{*}{UD v2.5} & $C/|Q|$ & 0.126 & [-0.08, 0.32] & 0.228 & 0.227\\
   & TTR & -0.048 & [-0.25, 0.16] & 0.644 & 0.075 \\[0.5em]
   \multirow{2}{*}{UD v2.6} & $C/|Q|$ & 0.136 & [-0.07, 0.33] & 0.202 & 0.248 \\
   & TTR & -0.038 & [-0.24, 0.17] & 0.724 & 0.064 \\
   \bottomrule
   \end{tabular}
\end{table}      

\begin{figure}[htbp!]
    \centering
    \includegraphics[width=0.8\linewidth]{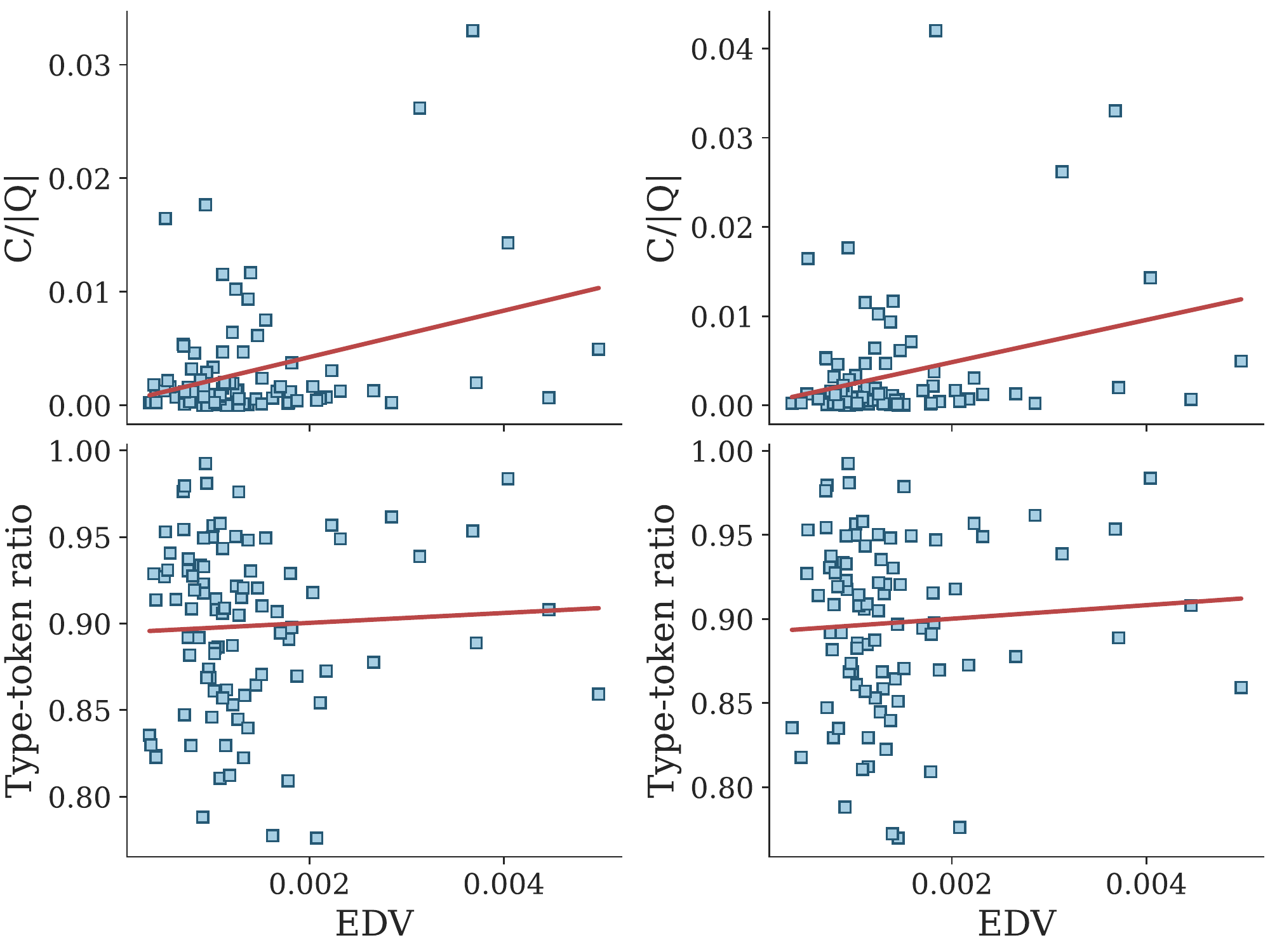}
    \caption{$C/|Q|$ (the normalized number of crossings found in a treebank) and type--token ratio of the training data found in UD v2.5 (column 1) and UD v2.6 (column 2) against the corresponding EDV for each treebank.}
    \label{fig:reviewer_b}
\end{figure}
\appendixsection{An aggregate measurement of morphological complexity}\label{appendix:complexity}
In this appendix, we give the details about the measurement we used for approximating morphologically complex subsets of the treebanks which were used in Section \ref{sec:mc}. It is an aggregate measurement, consisting of word entropy \citep{shannon1948a}, type--token ratio \cite{bentz2016a}, form to lemma ratio, form to inflected lemma ratio, and head part-of-speech entropy \citep{dehouck-denis-2018-framework}. These are normalized when needed such that 0 means no morphological complexity and 1 means the highest possible morphological complexity, so that we can simply take the mean measurement across all 5 metrics. 

\vspace{1em}\noindent\textbf{Normalized word entropy:~~}
Word entropy gives an indication as to how much information any given word has with a higher entropy resulting from a treebank having many forms. It is given by:
\begin{equation}
    H_{\textrm{word}} = - \sum_{v\in\mathcal{V}}p(v)\log_2p(v)
\end{equation}
where $\mathcal{V}$ is the vocab space in a given treebank, $v$ is a given word in that space, and $p(v)$ is the probability of that word occurring estimated by its frequency count \cite{shannon1948a}. The normalized word entropy, $H^*_\textrm{word}$, is obtained by dividing by the log of the magnitude of the vocab space: 
\begin{equation}
    H^*_\textrm{word} = \frac{H_\textrm{word}}{log_2|\mathcal{V}|}
\end{equation}
\noindent\textbf{type--token ratio:~~}
The type--token ratio gives an indication of the morphological production in a given treebank. It is given by:
\begin{equation}
    \mathit{TTR} = \frac{|\mathcal{V}|}{|T|}
\end{equation}
where $\mathcal{V}$ is the vocab space in a given treebank and $T$ is the number of tokens \cite{bentz2016a}. While this number isn't exactly bounded by 0 at the lower margin (it is bounded by 1 at the upper margin), when $T$ is suitably big, which is typically the case, the instance where $\mathcal{V}$ only consists of 1 type, $\mathit{TTR}$ tends to zero. However, this is clearly not a likely scenario in a treebank and so this inconsistency is not a worry in reality.

\vspace{1em}\noindent\textbf{Form to lemma ratio:~~}
The form to lemma ratio is similar to the type--token ratio but it more closely measures morphological production by honing in on lemmas having multiple forms rather than just looking at the more global measurement of production in TTR. It is given by: 
\begin{equation}
    F/L = \frac{1}{|\mathcal{L}|} \sum_{l\in\mathcal{L}}|\mathcal{F}_l|
\end{equation}
where $\mathcal{L}$ is the lemma vocab of a treebank, $l$ is a given lemma in the vocab, and $\mathcal{F}_l$ is the set of forms associated with $l$ \cite{dehouck-denis-2018-framework}.

As defined $F/L$ ranges from 1 to $|\mathcal{V}|$ (the absurd case of a singular lemma). By taking the reciprocal, we obtain a value that tends to zero in the absurd case and has an upper bound of 1. However, this gives us an inverse scale, i.e. a lower value means more morphology and a higher value less. Therefore we subtract the reciprocal of $F/L$ from 1:
\begin{equation}
    F/L^* = 1 - \frac{1}{F/L}
\end{equation}
\noindent\textbf{Inflected form to lemma ratio:~~}
This is the same as $F/L$ but for the case where a lemma is actually inflected, i.e. the case where the set of word forms associated with a given lemma is greater than 1. It is given by:
\begin{equation}
    F/iL = \frac{1}{|\mathcal{L}_2|}\sum_{l\in\mathcal{L}_2}|\mathcal{F}_l|
\end{equation}
where $\mathcal{L}_2$ is the subset of lemmas which have 2 or more forms associated with them in a treebank,   $l$ is a given lemma in that subset, and $\mathcal{F}_l$ is the set of forms associated with $l$ \cite{dehouck-denis-2018-framework}.  It is normalized in the same way as $F/L$:
\begin{equation}
    Fi/L^* = 1 - \frac{1}{F/iL}
\end{equation}
\noindent\textbf{Head part-of-speech entropy:~~} 
The head part-of-speech entropy (HPE) is the  measurement of morphology most related to parsing as it captures the morphosyntactic complexity found in a treebank. It is measured on the delexicalized version of the treebank, where the unit is a concatenation of a token's POS tag and morphological feature tags. The HPE of a treebank is an average over the HPE of each delexicalized word type:
\begin{equation}
    \mathit{HPE} = \frac{1}{|\mathcal{D}|} \sum_{d\in\mathcal{D}}  \mathit{HPE}_d 
\end{equation}
where:
\begin{equation}
    \mathit{HPE}_d = - \sum_{t\in\mathcal{T}_d} p(h_d^t)\log_2p(t_d^t)
\end{equation}
where $h_d^t$ denotes the head of $d$ having the POS tag $t$ from the tagset $\mathcal{T}_d$ (the set of tags that $d$ is headed by in the treebank) and $p(h_d^t)$ is the probability of this occuring based on frequency counts \cite{dehouck-denis-2018-framework}. As defined this gives a value that tends to zero when morphosyntactic complexity is prevalent and increases unbounded the less morphosyntactic complexity is present. In order to normalize this, we have to normalize $\mathit{HPE}_d$:
\begin{equation}
    \mathit{HPE}_d^* = \frac{\mathit{HPE}_d}{\log_2|\mathcal{T}_d|}
\end{equation}
such that the normalized head part-of-speech entropy is simply:
\begin{equation}
    \mathit{HPE}^* = 1 - \frac{1}{|\mathcal{D}|} \sum_{d\in\mathcal{D}}  \mathit{HPE}_d^* 
\end{equation}
Note that the sum over the normalized $HPE_d$ values is subtracted from 1 to invert the scale such that 0 denotes no morphosyntactic complexity and 1 the maximum. 

\vspace{1em}\noindent\textbf{Aggregate metric:~~}
The final metric we used is a simple unweighted average of the 5 normalized metrics described above:
\begin{equation}
    \mathit{MC} = \frac{(H_{\textrm{word}}^* + \mathit{TTR} + F/L^* + F/iL^* + \mathit{HPE}^*)}{5}
\end{equation}
Below are the lists with the treebanks considered \textit{morphologically complex} in UD v2.5 and v2.6, respectively. The code used for this is available at \url{https://github.com/markda/morphological-complexity}.

\vspace{1em}\noindent\textbf{List of morphologically complex treebanks in UD v2.5:~~}\vspace{1em}

\begin{footnotesize}
\begin{tabular}{lll}
Ancient Greek-PROIEL & Gothic-PROIEL & Persian-Seraji\\
Ancient Greek-Perseus & Greek-GDT & Polish-LFG\\
Armenian-ArmTDP & Hungarian-Szeged & Polish-PDB\\
Basque-BDT & Irish-IDT & Portuguese-GSD\\
Belarusian-HSE & Latin-ITTB & Romanian-Nonstandard\\
Bulgarian-BTB & Latin-PROIEL & Romanian-RRT\\
Croatian-SET & Latin-Perseus & Russian-GSD\\
Czech-CAC & Latvian-LVTB & Russian-SynTagRus\\
Czech-CLTT & Lithuanian-ALKSNIS & Russian-Taiga\\
Czech-FicTree & Lithuanian-HSE & Serbian-SET\\
Czech-PDT & Maltese-MUDT & Slovak-SNK\\
Estonian-EDT & Marathi-UFAL & Slovenian-SSJ\\
Estonian-EWT & North Sami-Giella & Slovenian-SST\\
Finnish-FTB & Old Church Slavonic-PROIEL & Tamil-TTB\\
Finnish-TDT & Old French-SRCMF & Telugu-MTG\\
German-HDT & Old Russian-TOROT & Turkish-IMST\\
\end{tabular}
\end{footnotesize}

\vspace{1em}\noindent\textbf{List of morphologically complex treebanks in UD v2.6:~~}\vspace{1em}

\begin{footnotesize}
\begin{tabular}{lll}
Ancient Greek-PROIEL & Greek-GDT & Old Russian-TOROT\\
Ancient Greek-Perseus & Hungarian-Szeged & Persian-Seraji\\
Armenian-ArmTDP & Irish-IDT & Polish-LFG\\
Basque-BDT & Latin-ITTB & Polish-PDB\\
Belarusian-HSE & Latin-PROIEL & Portuguese-GSD\\
Bulgarian-BTB & Latin-Perseus & Romanian-RRT\\
Croatian-SET & Latvian-LVTB & Russian-GSD\\
Czech-CAC & Lithuanian-ALKSNIS & Russian-Taiga\\
Czech-CLTT & Lithuanian-HSE & Sanskrit-Vedic\\
Czech-FicTree & Maltese-MUDT & Serbian-SET\\
Estonian-EDT & Marathi-UFAL & Slovak-SNK\\
Estonian-EWT & North Sami-Giella & Slovenian-SSJ\\
Finnish-FTB & Old Church Slavonic-PROIEL & Slovenian-SST\\
Finnish-TDT & Old French-SRCMF & Tamil-TTB\\
Gothic-PROIEL & Old Russian-RNC & Telugu-MTG\\
\end{tabular}
\end{footnotesize}
\appendixsection{Variance in EDV for different sizes of training data}
\renewcommand\thefigure{D.\arabic{figure}} 
\renewcommand\thetable{D.\arabic{table}} 
\begin{table}[b!]
    \footnotesize
    \centering
    \caption{Standard deviation is reported with respective means in the form mean (standard deviation). Each value $x$ (other than full training count) in the table corresponds to $x\times10^{-4}$. Values are given for sample sizes of 2K, 4K, 6K, and 8K training sentences. The final column "Full training" gives the total number of sentences in the original training data. 
    Standard deviation ranges from $3.7\%$ to $17.1\%$ of respective means ($5.6\%$ on average).}
    \label{tab:size_variance_table}
    \begin{tabular}{lcrccr}
    \toprule
    & \multicolumn{4}{c}{EDV  --- mean$\times10^{-4}$ (standard deviation$\times10^{-4}$)} \\
    & 2K Sample & \multicolumn{1}{c}{4k Sample} & 6k Sample & \multicolumn{1}{c}{8k Sample} & \multicolumn{1}{r}{Full training}\\
    \midrule
\textbf{Czech-PDT} & $3.8\;(0.7)$ & $3.4\;(0.5)$ & $3.2\;(0.3)$ & $3.3\;(0.4)$ & 68,495 \\
\textbf{Estonian-EDT} & $5.8\;(0.7)$ & $5.9\;(0.6)$ & $5.6\;(0.4)$ & $5.4\;(0.3)$ & 24,633 \\
\textbf{German-HDT} & $4.1\;(0.7)$ & $3.6\;(0.5)$ & $3.3\;(0.4)$ & $3.3\;(0.3)$ & 153,035 \\
\textbf{Japanese-BCCWJ} & $9.9\;(0.9)$ & $10.0\;(0.7)$ & $9.6\;(0.5)$ & $9.8\;(0.5)$ & 40,740 \\
\textbf{Korean-Kaist} & $5.4\;(0.5)$ & $5.3\;(0.5)$ & $5.2\;(0.3)$ & $5.0\;(0.2)$ & 23,010 \\
\textbf{Persian-PerDT} & $6.2\;(0.8)$ & $5.8\;(0.5)$ & $5.8\;(0.5)$ & $5.9\;(0.3)$ & 26,196 \\
\bottomrule
\end{tabular}
\end{table}
We offer a small analysis on how much variance we observe in the same treebank when sampling smaller amounts of training data, as it is possible that evaluation undertaken in this work would only hold true for these very specific splits. Although the sampling evaluation makes this unlikely.

We selected treebanks from UD v2.7 which had at least 20,000 training instances, so that samples could be sufficiently different. We opted for Czech-PDT, Estonian-EDT, German-HDT, Japanese-BCCWJ, Korean-Kaist, and Persian-PerDT as this offered us the best spread across different languages and language families from the largest treebanks. We then sampled different amounts of training data for each of these treebanks. We sampled 2,000; 4,000; 6,000; and 8,000 training samples. For each training size, we sampled 20 unique sets of training instances.

\begin{figure}[tb!]
    \centering
    \includegraphics[width=0.9\linewidth]{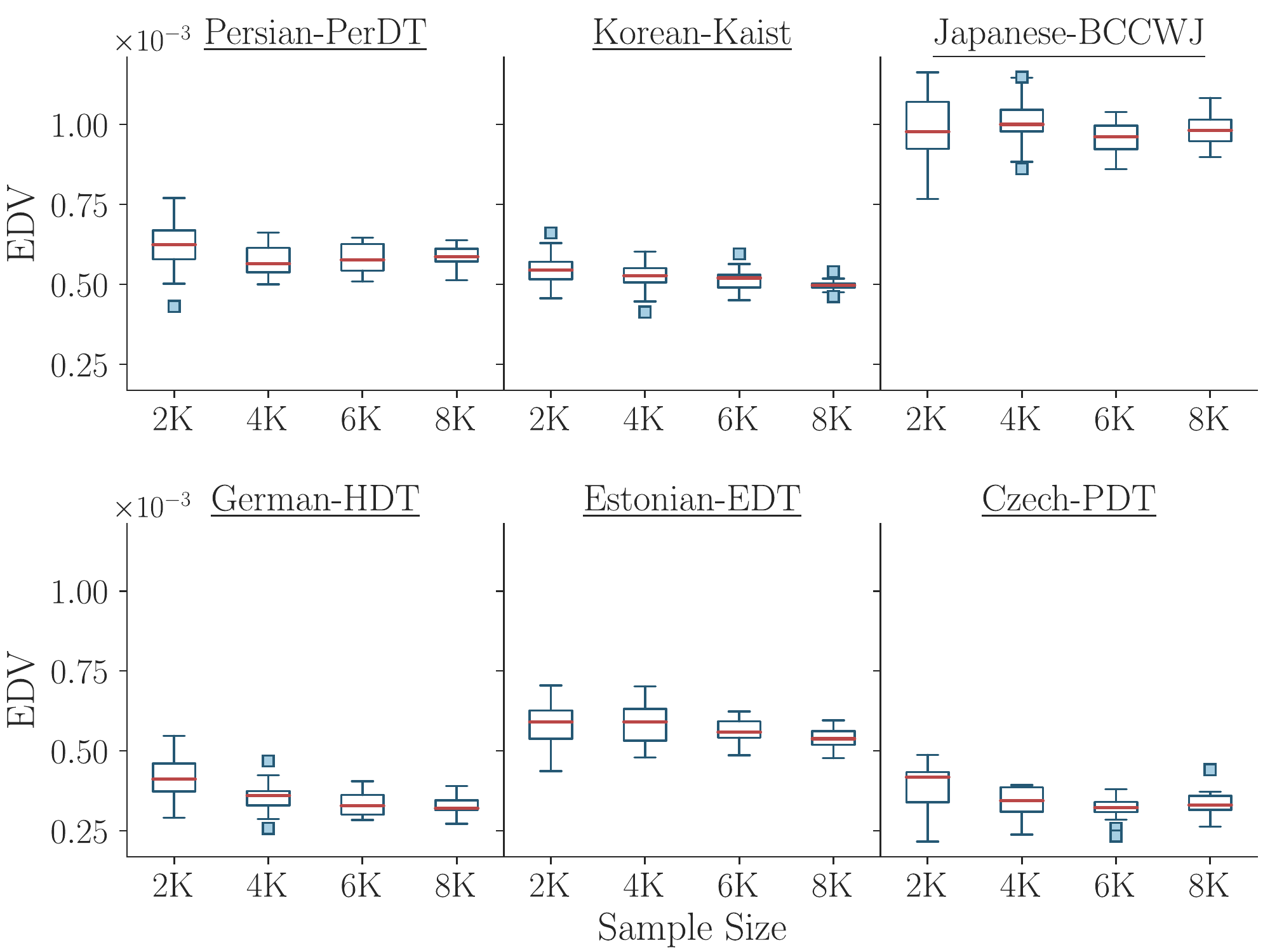}
    \caption{Distributions of EDV for different treebanks with varying sample sizes. Smaller sample sizes exhibit greater variance than larger sample sizes, but not to such a degree that the values measured for EDV for different languages change how they compare to those of other treebanks.}
    \label{fig:size_variance}
\end{figure}

Figure \ref{fig:size_variance} shows the distributions of EDV values split across each treebank for the different training size. The first thing that is clear is that across different sample sizes, the differences observed across treebanks are fairly stable. However, for most treebanks the variance in EDV decreases as the sample size increases. This is expected and the variance observed for the smallest sample size is still not particularly high. Table \ref{tab:size_variance_table} gives the corresponding mean and standard deviation from the values used in Figure \ref{fig:size_variance}. This further corroborates that the variance in EDV for the smaller sample sizes are not problematically high, with the standard deviation averaging at 5.6\% of their respective means.

\clearpage

\section*{Acknowledgments}
This work has received funding from the European Research Council (ERC), under the European Union's Horizon 2020 research and innovation program (FASTPARSE, grant agreement No 714150), from ERDF/MICINN-AEI (ANSWER-ASAP, TIN2017-85160-C2-1-R; SCANNER-UDC, PID2020-113230RB-C21), from Xunta de Galicia (ED431C 2020/11), and from Centro de Investigación de Galicia ``CITIC'', funded by Xunta de Galicia and the European Union (ERDF - Galicia 2014-2020 Program), by grant ED431G 2019/01.
\starttwocolumn
\bibliography{clv3}

\end{document}